%% file: main.tex
\documentclass[postscript]{article}

\usepackage[preprint]{neurips_2020}
\usepackage[dvipsnames]{xcolor}

\definecolor{darkraspberry}{rgb}{0.53, 0.15, 0.34}
\usepackage{amsmath}

\usepackage{setspace}

\usepackage{xcolor}
\usepackage[postscript, cjkjis]{ucs}
\usepackage{pifont}
\usepackage[whole]{bxcjkjatype}
\usepackage{fdsymbol}
\usepackage{newunicodechar}
\newunicodechar{◼}{$\mdblksquare$}

\newcommand\Tstrut{\rule{0pt}{2.6ex}}

\input{highlight}

\usepackage[T1]{fontenc}    
\usepackage[utf8]{inputenc} 
\usepackage[linktocpage=true]{hyperref}
\usepackage{url}            
\usepackage{booktabs}       
\usepackage{amsfonts}       
\usepackage{nicefrac}       
\usepackage{microtype}      
\usepackage{wrapfig}
\usepackage{todonotes}
\usepackage{graphicx, caption, subcaption, xcolor, multirow, array, enumitem}
\newcolumntype{P}[1]{>{\centering\arraybackslash}p{#1}}

\newcommand{\comment}[1]{}

\hypersetup{
  colorlinks   = true, 
  urlcolor     = BlueViolet, 
  linkcolor    = Fuchsia,
  citecolor   = blue 
}

\title{Efficient Training of Language Models to \\ Fill in the Middle}

\author{
Mohammad Bavarian \thanks{Equal contribution, order randomized. Correspondence to: \href{mailto:mobav@openai.com}{\nolinkurl{mobav@openai.com}}, \href{mailto:heewoo@openai.com}{\nolinkurl{heewoo@openai.com}}.}
\And
Heewoo Jun\footnotemark[1]
\And
Nikolas Tezak
\AND
John Schulman
\And 
Christine McLeavey
\And
Jerry Tworek
 \And
Mark Chen
\AND
{\normalfont\large OpenAI}
}

\begin{document}

\maketitle

\begin{abstract}
We show that autoregressive language models can learn to infill text after we apply a straightforward transformation to the dataset, which simply moves a span of text from the middle of a document to its end. While this data augmentation has garnered much interest in recent years, we provide extensive evidence that training models with a large fraction of data transformed in this way does not harm the original left-to-right generative capability, as measured by perplexity and sampling evaluations across a wide range of scales. Given the usefulness, simplicity, and efficiency of training models to fill-in-the-middle (FIM), we suggest that future autoregressive language models be trained with FIM by default. To this end, we run a series of ablations on key hyperparameters, such as the data transformation frequency, the structure of the transformation, and the method of selecting the infill span. We use these ablations to prescribe strong default settings and best practices to train FIM models. We have released our best infilling model trained with best practices in our API, and release our infilling benchmarks to aid future research.
\clearpage
\end{abstract}

\begingroup
\small
\hypersetup{linkcolor=black, colorlinks=True}
\tableofcontents
\endgroup

\begin{figure}[ht!]
\centering
\includegraphics[width=\textwidth]{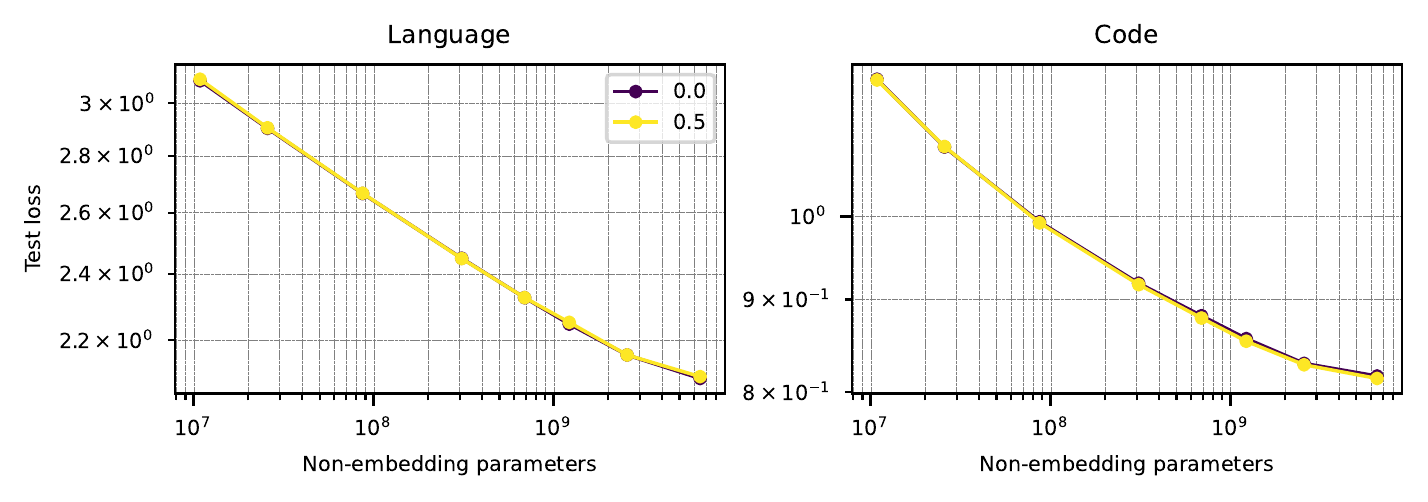}
\caption{FIM can be learned for free. We pretrain language models with 50\% and 0\% FIM rates on two domains, natural language and code, and evaluate the test loss of all the final snapshots. All models are trained on 100B tokens of data. We observe that joint FIM training incurs no cost as the original left-to-right loss trend remains the same even though FIM models see the original data only 50\% of the time and the models are learning a new capability. See Figure \ref{fig:no-harm-perf} for more evidence for the FIM-for-free property.}
\label{fig:no-harm-perp}
\end{figure}

\section{Introduction}\label{sec:intro}

Following the introduction of the Transformer \citep{transformer}, large language models (LLMs) trained on diverse Internet scale datasets have achieved remarkable success. These models are capable of producing coherent and sensible completions given a natural language prompt, and they achieve state-of-the-art performance in many benchmarks including reading comprehension, question answering, logical inference, and common sense reasoning.

There are several broad classes of transformer based language models: encoder-only models like BERT \citep{bert} are typically trained with a masked language modeling objective, and encoder-decoder models like T5 \citep{T5} are typically trained with a span prediction objective \citep{mass}. Finally, causal decoder-based language models, like the GPT model series \citep{gpt1, gpt2, gpt3}, are trained using the left-to-right next token prediction objective. The largest and most capable generative language models today, such as GPT-3, Codex, LaMDA, GLaM, PaLM, Gopher, Jurassic-1, and Chinchilla \citep{gpt3, codex, lamda, glam, palm, gopher, jur, chinchi}, belong to the latter class of models. The overwhelming popularity of the causal decoder-based models at the largest scale is due to their superiority in open-ended text generation, in-context learning (using few-shot priming), pretraining computational efficiency \citep{huggingface}, and to some extent historical precedence in successful scale-ups \citep{gpt3}. These models are also architecturally simpler and generally more effective without task specific finetuning, making them more attractive for inference and deployment.

All model classes are limited when it comes to infilling, where the model is tasked with generating text at a specific location within a prompt, while conditioning on both a prefix and a suffix. Left-to-right models can only condition on the prefix. While encoder-only and encoder-decoder models are capable of conditioning on suffixes, the lengths of infill regions seen at training time are typically much shorter than what is useful in practice. This is unfortunate because infilling naturally arises in applications where there is context both before and after the point of generation. For example, in creating a coding assistant, infilling can be used for docstring generation, import statement generation, or for completing a partially written function.

Our goal in this work is to address this limitation by adding fill-in-the-middle (FIM) capability to causal decoder-based language models which are currently the most dominant paradigm for large scale language modelling \citep{gpt3, chinchi, palm}. We show that with a simple modification to training data and without changing the model architecture, causal decoder-based autoregressive (AR) language models can learn infilling \emph{without compromising their normal left-to-right generative capability}.

\begin{figure}[ht!]
\centering
\includegraphics[width=\textwidth]{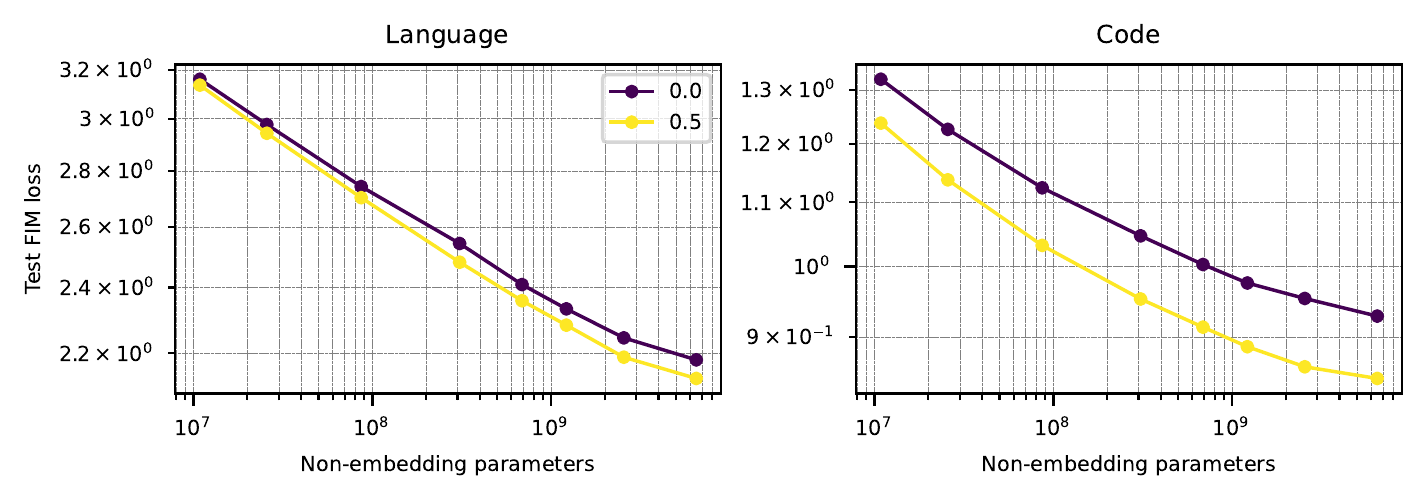}
\caption{Evaluation of infilling capabilities of the same model scans from Figure \ref{fig:no-harm-perp} using FIM test losses. Models trained with FIM (yellow) obtain lower FIM test loss than baseline (purple) AR models. This shows that the FIM models are indeed learning to condition on the suffix while predicting the middle section allowing them to achieve lower test loss on FIM test set. Figures \ref{fig:no-harm-perp} and \ref{fig:fim_loss} together indicate that FIM models can be considered strictly better than AR models as they achieve the same left-to-right autoregressive loss but lower FIM loss.}
\label{fig:fim_loss}
\end{figure}

The key to our approach, described in Section \ref{sec:fim_training}, is a transformation applied to a fraction of our dataset, in which we split documents into three pieces at random and move the middle piece to the end:
\[  \text{document} \rightarrow (\text{prefix}, \text{middle}, \text{suffix}) \rightarrow (\text{prefix}, \text{suffix}, \text{middle}) \]
We then concatenate the three pieces using sentinel tokens. This is similar to the procedure used in \citep{donahue, cm3, incoder}.

Compared to prior work, our work emphasizes the computational efficiency of training FIM models. This emphasis is important given the increased interest in training very large language models, which are very expensive to train and have a substantial energy footprint. In general, when adding a new objective or capability to language models, we believe the most critical question is the effect on the existing capabilities and the computational efficiency trade-offs. 

Unlike most cases where we jointly train on multiple objectives and datasets, we show that models trained jointly on a mixture of FIM transformed data and ordinary left-to-right data achieve the same left-to-right capability while learning how to fill-in-the-middle. We call this the FIM-for-free property.

In what follows, we use the term FIM model to refer to any model trained on a mixture of FIM transformed and normal left-to-right data. We refer to models trained without any FIM data (i.e.~0\% FIM rate) as AR models. 

 \subsection{Our contributions}\label{sec:intro:contrib}
  Our central contributions in this paper are as follows:

 \begin{itemize}
     \item \textbf{FIM-for-free property}: We perform an extensive scaling study by training a suite of 8 models, with and without FIM, and show that FIM can be learned without compromising the left-to-right capability in pretraining. We examine this claim in both code and language, using both perplexity and sampling-based benchmarks. 
    \item \textbf{Best practices for FIM in pretraining}: We clarify the effects of many hyperparameters related to training FIM models using comprehensive ablations. In particular, we study the FIM rate (the probability at which FIM transformation is applied to the data), different variants of FIM transformation, and the choice of middle span.
    \item \textbf{Finetuning inefficiency}: An alternative to training FIM models from scratch is to learn this capability by finetuning existing language models. We show that finetuning with FIM is computationally inefficient. While FIM can be learned for free during pretraining, learning FIM during finetuning requires a significant amount of additional compute to reach similar levels of performance as pretraining.
    \item \textbf{New infilling benchmarks.} In order to study the generative capabilities of our models, we need to evaluate the correctness of free-form generated samples. For this, we focus on code where we can use unit tests to evaluate the correctness of long FIM samples. In particular, we use the single-line and multi-line infilling benchmarks introduced by \citep{incoder} by removing non-empty lines of canonical solutions of HumanEval \citep{codex}. However since line-based evaluations do not capture all the use cases of FIM, we create two new benchmarks called \emph{random span infilling} and \emph{random span infilling light}. 
    We discuss these benchmarks and our evaluation methodology more generally in Section \ref{sec:setup:evals}. 
    \item \textbf{Need for sampling evaluations}.  In  Sections \ref{sec:pretraining:fim-rate}, \ref{sec:pretraining:context}, and Appendix \ref{appendix:more_fim_rate}, we find that changing various hyperparameters in FIM training often leads to negligible differences in FIM test losses but large differences in sampling based benchmarks. Not only are sampling benchmarks closer to real use cases, but they also appear to be able to tease apart gains that can be missed using test losses. This is an important finding since often scaling laws analysis relies just on test losses, which we find are misleading if not augmented with other evaluations.
 \end{itemize}
 It is interesting to contrast the first and third bullet points above. The first states that learning FIM in pretraining is free while leaving it to finetuning is surprisingly costly. We discuss potential explanations for this finding in Section \ref{sec:discussion}. To establish the FIM-for-free property, we perform an ablation study on both code and language across a range of scales. We train 8 models from 50M to 6.9B parameters, both with and without FIM, and compare the performance across a variety of autoregressive benchmarks. In particular, we train 16 models on code for 100B tokens and another 16 models on natural language for 100B tokens. The comparison of these models in terms of normal autoregressive left-to-right language modeling test loss is presented in Figure \ref{fig:no-harm-perp}. In both domains, FIM models achieve similar AR test loss as the non-FIM models.  

We provide more evidence for the FIM-for-free property by comparing FIM and AR models on non-loss based benchmarks in Section \ref{sec:pretraining}. Moreover, we see in Section \ref{sec:pretraining:fim-rate} that there is a \emph{stronger form of the FIM-for-free property}. Not only there is no hit in autoregressive capabilities from FIM training on the final checkpoints, the same also holds throughout training. This is evidenced by the matching learning curves between AR and FIM models in left-to-right loss and HumanEval evaluations in Figures \ref{fig:pretraining:fim-rate1} and \ref{fig:pretraining:fim-rate2}.

Beside studying the effect of FIM training on the left-to-right capability, it is also important to show that the models are in fact learning to infill from FIM training. Figure \ref{fig:fim_loss} provides evidence for this in the context of FIM test losses. We study the infilling capabilities of our models more extensively in Section \ref{sec:pretraining} and Appendix \ref{appendix:qual_eval}.

\section{Evaluation}
\label{sec:setup:evals}

We use both AR and FIM evaluation benchmarks to analyze the capabilities of our models. Vanilla AR evaluation is important for quantifying the impact of FIM training on left-to-right capabilities and allows us to demonstrate the FIM-for-free property from Section \ref{sec:intro:contrib}. FIM evaluation is important for understanding the effect of different hyperparameters on FIM training and to understand the scaling trends.

Throughout the paper, we use the terms AR and left-to-right interchangeably. AR loss refers to the cross entropy loss on normal left-to-right data and FIM loss as the loss on 100\% FIM transformed data. All test losses are in nats per token unit. In all sampling-based benchmarks, we use nucleus sampling \citep{nucleus} with a nucleus parameter of 0.95. 

\subsection{Autoregressive evaluation}
\label{sec:setup:eval:ar}

For all domains, we evaluate test losses in the canonical autoregressive order to show that the learning curves and scaling trends remain the same even with FIM augmentation. Beside test losses,  we evaluate on standard benchmarks to demonstrate that the model's capabilities are unaffected by FIM training. For natural language, we use PIQA \citep{piqa}, Winograd \citep{winograd}, WinoGrande \citep{winogrande} for common sense reasoning, DROP \citep{drop} and QuAC \citep{quac} for reading comprehension, and HellaSwag \citep{hellaswag}, LAMBADA \citep{lambada}, StoryCloze \citep{storycloze} for completion tasks. All benchmarks with the exception of DROP and QuAC
are evaluated with few-shot prompting. For code, we measure the pass rates on HumanEval \citep{codex}.

\subsection{Infilling evaluation}
\label{sec:setup:eval:fim}

To create FIM tests, we apply the FIM transformation to the examples from the AR test sets with a FIM rate of 100\%. Using the same underlying examples in FIM and AR test sets allows us to compare FIM and AR test losses. Additionally, we create a masked version of these test sets where we only measure the loss on the middle span tokens. The latter test sets are used to measure $P(\text{middle}|\text{prefix}, \text{suffix})$ for FIM models and $P(\text{middle}|\text{prefix})$ for AR models allowing us to investigate the amount of information FIM models gain by being able to condition on the suffix.  

For generative infilling capabilities, we focus on code since we are interested in free-form generation in contrast to single or few token generations common in cloze-style natural language benchmarks. The advantage of working with code is that we can use test suites to evaluate the correctness of samples in our tasks even when evaluating long samples from open-ended generations.

All the sampling based infilling benchmarks we use are partial function completions tasks created by  removing middle spans from the canonical solutions of  HumanEval \citep{codex}. In particular, we use the single-line and multi-line infilling benchmarks proposed by \citep{incoder} where different spans of non-empty lines in the canonical solutions of HumanEval are turned into a FIM task. In addition, we create a new benchmark called random span infilling\footnote{Released at \url{https://www.github.com/openai/human-eval-infilling}}, where for each HumanEval problem, we create infilling tasks by selecting the middle span from the canonical solution uniformly at random. We show an example of such a task below where the model must predict the highlighted section (or an alternative completion accomplishing the same goal). We refer to Appendix \ref{appendix:random-span-infilling} for more details.

\vspace{-6pt}
\begin{center}
\begin{minipage}{5in}
\begin{lstlisting}[style=python]
def unique(l: list):
    """Return sorted unique elements in a list
    >>> unique([5, 3, 5, 2, 3, 3, 9, 0, 123])
    [0, 2, 3, 5, 9, 123]
    """
    ret`urn sorted(list(set(l`)))
\end{lstlisting}
\end{minipage}
\end{center}
\vspace{-6pt}

The single-line, multi-line, and random span infilling  together constitute our infilling benchmark suite. These benchmarks have  1033, 5815, and 1640 tasks, respectively. We note that this is much larger than the number of tasks in the original HumanEval dataset (164 tasks), which reduces variance in our evaluations. Still, we take at least 100 to 200 samples per task to further reduce variance when evaluating these benchmarks on the final snapshots of our models. We also use random span infilling \emph{light}, a smaller version of random span infilling, with only one random FIM task per HumanEval problem and just 164 tasks, to  track the infilling capability trends during training.

In Section \ref{sec:fim_training}, we find that FIM can be prepared in two different ways denoted as PSM and SPM. We report just the SPM infilling results for brevity, except in cases when the use of PSM changes the conclusions.

\section{FIM training and inference}
\label{sec:fim_training}

We implement FIM using a random transformation applied to our dataset. We experiment with two different implementations: document level and context level.
The difference between the two is at which stage of the data loading pipeline the FIM transformation occurs.
This choice naturally arises because a long document can be broken into many contexts, or a context can contain multiple documents when the documents are small.
We first describe the document-level case and then describe the changes required to implement context-level FIM in Section \ref{sec:contex_impl}. 

In document-level FIM, with a certain probability $p$ called the \textbf{FIM rate} (we use $p=0.5$ for our main suite of models), we cut each document into three parts: prefix, middle, and suffix. We perform this split prior to tokenization, when the document is still a sequence of characters. We split uniformly at random, which means the lengths of prefix, middle, and suffix are each 1/3 of the full document in expectation.

We then encode each of the three sections separately and prepend sentinel tokens to the beginning of each section. We denote these sentinel tokens by $\textsc{<pre>}$, $\textsc{<mid>}$, and $\textsc{<suf>}$. Finally we concatenate all these sections in the order prefix, suffix, and middle along with their sentinel tokens to form the tokenized version of the FIM document,

\[
  \textsc{<pre>} \circ \text{Enc}(\text{prefix}) \circ \textsc{<suf>}\circ \text{Enc}(\text{suffix}) \circ \textsc{<mid>}\circ \text{Enc}(\text{middle}),   \tag{PSM}
\]
where $\circ$ denotes concatenation. The different documents, whether FIM or AR, then are concatenated with $\textsc{<eot>}$ and are given to the model during training. 
We reiterate
that we keep the loss on all three sections prefix, middle, and suffix, so FIM training does not cause a decrease in the autoregressive learning signal. Preliminary experiments, although not reported here, suggest that this choice is crucial for the FIM-for-free property to hold. This property does not change whether the sentinels are masked or not; however, it is important to always train on the \textsc{<eot>} tokens as it signals a successful join to the suffix.

For inference, we encode the given prefix and suffix and prompt the model with 
\[
 \textsc{<pre>} \circ \text{Enc}(\text{prefix}) \circ \textsc{<suf>}\circ \text{Enc}(\text{suffix}) \circ \textsc{<mid>}.\footnote{It is worth noting that prepending this prompt with $\textsc{<eot>}$  leads to a slight improved performance, and we do so when evaluating our models in sampling benchmarks.}   \tag{PSM inference}
\]
We continue sampling from the model until it generates the \textsc{<eot>} token which is how the model communicates it has connected the prefix and the suffix.

If the model fails to generate an $\textsc{<eot>}$ token within a reasonable allotted inference token budget, it is often a sign the model is having a difficult time connecting the prefix and the suffix, and the resulting samples often will be of worse quality, which motivates the procedure of EOT aware best-of-n sampling. See Appendix \ref{appendix:qual_eval} for more discussion.

\subsection{SPM mode}
\label{sec:fim:spm}

We also introduce a variant of the above procedure where we swap the order of prefix and suffix, called SPM, to emphasize the changing of the order to suffix, prefix, and middle. Our main motivation for introducing SPM is improved key-value caching during inference.
The reason for this advantage is that with SPM, appending tokens to the prefix no longer invalidates 
the keys and values computed in the suffix section. Note that superiority of SPM caching is not universal and may depend on the applications. In particular, in the SPM mode, minor changes to the suffix will invalidate the cache for prefix, but we expect changes to the suffix to be rarer than changes in prefix in real workloads. Interestingly, we find in Section \ref{sec:pretraining:psm-spm} beside the caching advantages, SPM  in fact has also a slight edge over PSM in the infilling benchmarks.

In our main runs, we apply the FIM transformation with 50\% probability in PSM mode and with 50\% probability in SPM mode, so the model is able to handle both types of formatting in inference. In other words, each mode inherits half of the total FIM rate $p$. We ablate this choice of joint training on PSM and SPM and compare with pure PSM and SPM runs. The results in Table \ref{tab:pretraining:psm-spm} show the efficacy of this choice. 

Even though the idea of SPM mode is simple, there are some subtleties with the placement of sentinel tokens in SPM which are especially important when training jointly on SPM and PSM. We describe these subtleties in Appendix \ref{appendix:spm_details}.
 
\subsection{Context-level FIM}\label{sec:contex_impl}

In language model training,  documents are often joined with a boundary token, referred to as $\textsc{<eot>}$, and are then chunked to the model context length.
When  applying FIM to long documents, this operation can result in fragmented FIM data where the entire prefix or suffix could get cut out of the context during chunking. To address this issue, we can apply FIM after the chunking step. A context slice may have multiple documents in them joined with the $\textsc{<eot>}$ boundary token. So, we split based on $\textsc{<eot>}$, turn some of the documents into FIM examples with probability given by the FIM rate, and join the examples again with $\textsc{<eot>}$. The resulting slice is then trimmed to the model context length. We refer to Appendix \ref{appendix:impl_details} for more details of FIM transformation.
In Section \ref{sec:pretraining:context}, we show this technique can boost performance relative to document-level FIM, and adopt context-level FIM in all our main FIM runs in this work.

\section{Pretraining results}
\label{sec:pretraining}

In Section \ref{sec:intro:contrib}, we discussed the FIM-for-free property which states that FIM can be learned without any impact to the left-to-right capability. We start this section by presenting more evidence for this result. Next, we study the hyperparameters of FIM training including the FIM rate, PSM vs SPM vs joint training, context vs document-level FIM, and the choice of middle span. Although FIM is free from the point of view of AR capability, the FIM capabilities themselves depend strongly on these hyperparameters. We study these choices in the code domain, where we can measure the correctness of generated samples using unit tests.

The models, unless otherwise stated, are trained with fixed horizon of 100B tokens. For our main scans we use all the 8 models described in Appendix \ref{appendix:arch}. For more extensive scans, e.g.~Sections \ref{sec:pretraining:fim-rate}, \ref{sec:pretraining:context}, and Appendix \ref{appendix:more_fim_rate}, we use a subset of the models trained with a shorter horizon to limit the compute costs.

\begin{figure}[ht!]
\centering
\begin{subfigure}[b]{\textwidth}
\centering
\includegraphics[width=\textwidth]{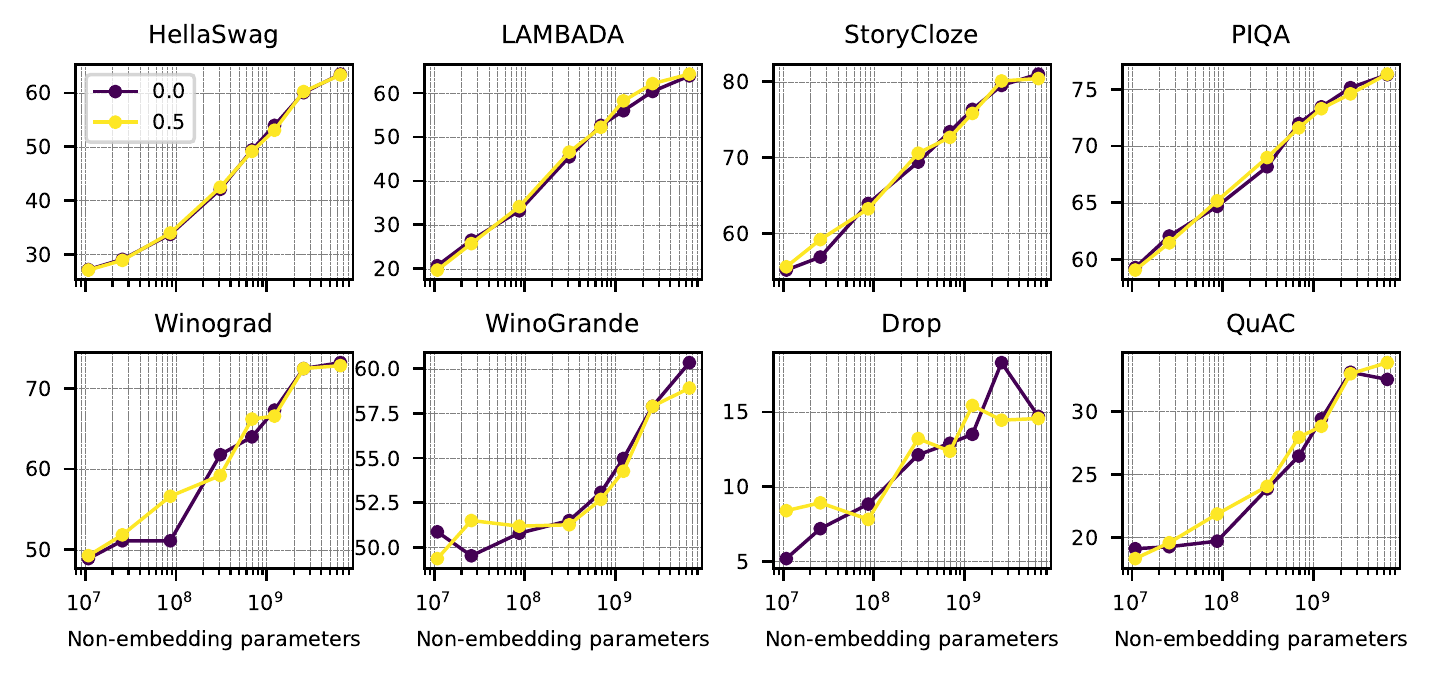}
\caption{Comparison of natural language results. We report F1 for Drop and QuAC and accuracy for the rest.}
\label{fig:no-harm-perf:lang}
\end{subfigure}
\begin{subfigure}[b]{\textwidth}
\centering
\includegraphics[width=\textwidth]{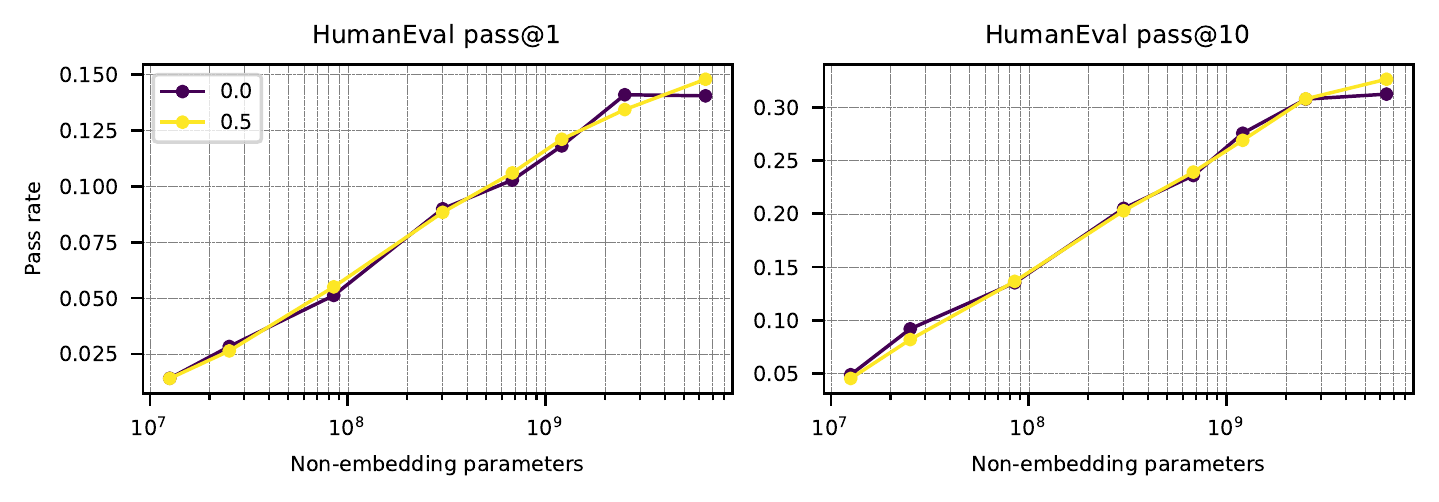}
\caption{Comparison of code results. We use temperature 0.8 and 400 samples per task for both pass@k.}
\label{fig:no-harm-perf:code}
\end{subfigure}
\caption{Comparison of performance on standard benchmarks for the natural language (top) and code (bottom) domains. Joint training of next-token prediction and FIM allows the model to learn the new infilling task without affecting the original capabilities. This provides further evidence for FIM-for-free property.}
\label{fig:no-harm-perf}
\end{figure}
\subsection{Evaluation of left-to-right capabilities in downstream benchmarks}
\label{sec:pretraining:eval_left_to_right}
We train a series of models from 50M to 6.9B parameters  from scratch with and without 50\% FIM augmentation on natural language and code domains. Figure \ref{fig:no-harm-perp} shows that the left-to-right test loss is unaffected even though FIM models see the data in its original form half the time, and are simultaneously learning a new skill. 

However, as we demonstrate below (see Sections \ref{sec:pretraining:fim-rate} and \ref{sec:pretraining:context}) it is often not sufficient to just consider test loss. So to strengthen the above results, we evaluate our models on a suite of standard downstream benchmarks, the result of which is presented in Figure \ref{fig:no-harm-perf}. We again find that joint FIM pretraining does not result in any degradation in standard AR benchmarks as the performance matches within error for both natural language and code.

\subsection{FIM rate}
\label{sec:pretraining:fim-rate}

\begin{figure}[ht!]
\centering
\includegraphics[width=\textwidth]{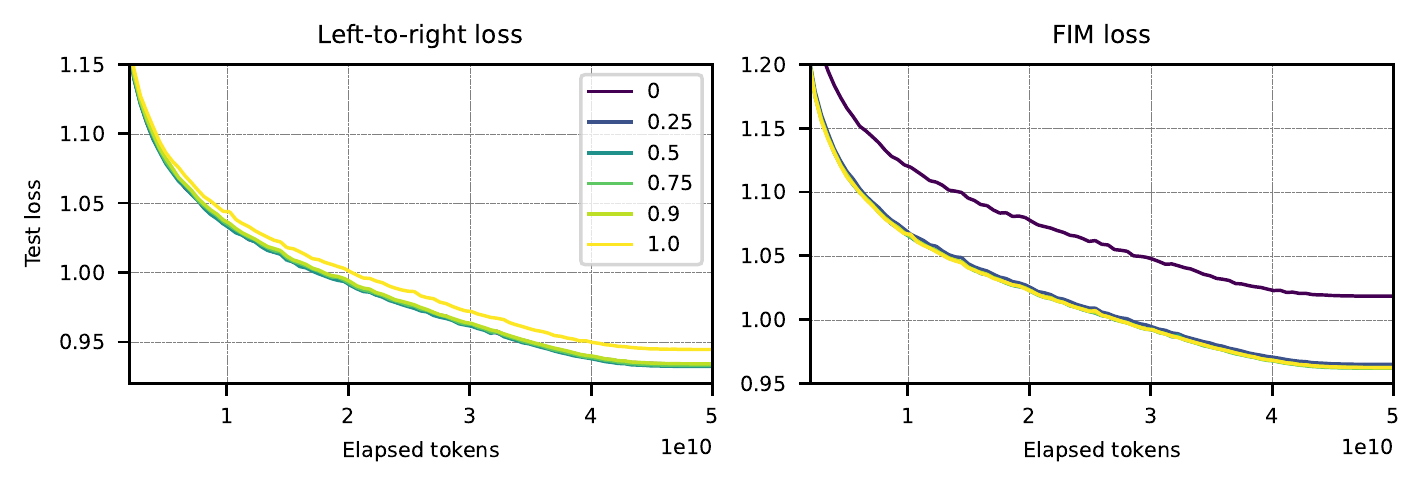}
\caption{Comparison of the learning curves of large (see Table \ref{tab:arch}) models trained with different FIM rates for 50B tokens. A FIM rate even up to 90\% does not have a noticeable effect on left-to-right test loss; however, at a FIM rate of 100\% there is degradation. We can also see the \emph{stronger FIM property} in the left figure: all runs with FIM rates less than 100\% follow very closely to the original left-to-right test loss.  }
\label{fig:pretraining:fim-rate1}
\end{figure}

\begin{figure}[ht!]
\centering
\includegraphics[width=\textwidth]{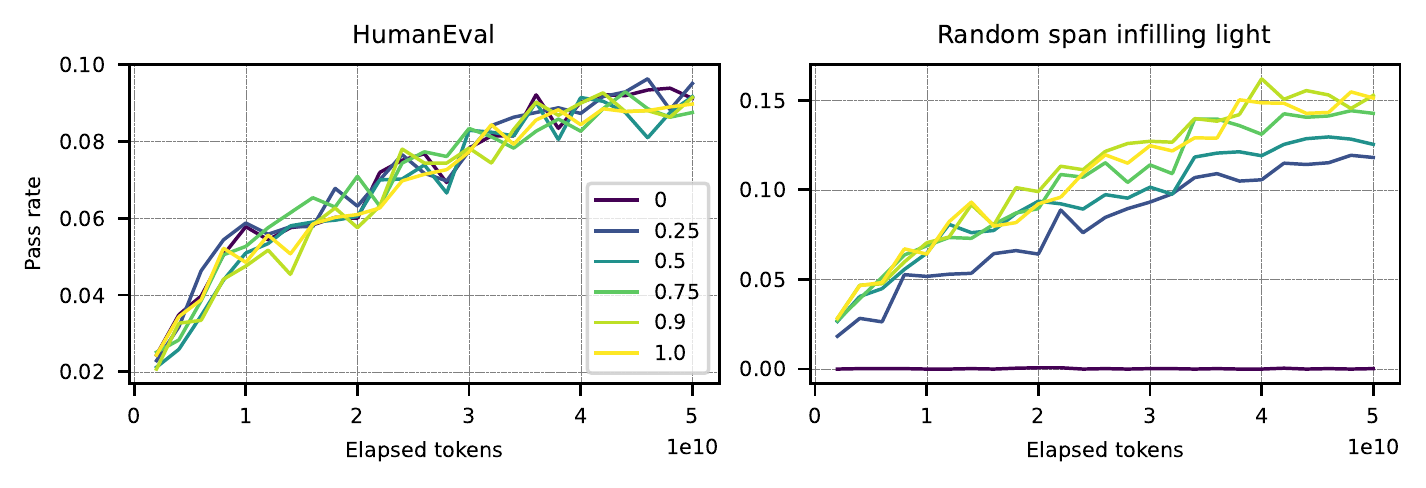}
\caption{In-run evaluation of coding benchmarks with temperature 0.8 and 25 samples per task. Using higher FIM rates do not have a noticeable effect on HumanEval performance. A higher FIM rate shows stronger infilling capabilities on the light random span infilling benchmark.}\label{fig:pretraining:fim-rate2}
\end{figure}

From Figures \ref{fig:no-harm-perp} and \ref{fig:no-harm-perf}, we see that a FIM rate of 50\% incurs no performance hit in the left-to-right capabilities. This naturally raises several questions:
\begin{itemize}
    \item Does FIM-for-free still hold even at higher FIM rates? If so, how high can we increase the FIM rate without degrading the left-to-right capabilities?
    \item Does using a higher FIM rate lead to stronger FIM capabilities? Or does the benefit saturate after a threshold?
\end{itemize}
In this section, we ablate the FIM rate to answer these questions. We train 6 large models (see Table \ref{tab:arch}) 
with FIM rates (0, 0.25, 0.5, 0.75, 0.9, 1.0) for 50B tokens. The results are presented in Figures \ref{fig:pretraining:fim-rate1} and \ref{fig:pretraining:fim-rate2}. The left plot in Figure \ref{fig:pretraining:fim-rate1} provides evidence that a FIM rate even up to 90\% does not cause any degradation in left-to-right capabilities.
However, there is a clear sign of degradation in ordinary AR test loss with 100\% FIM rate. For HumanEval, the left plot in Figure \ref{fig:pretraining:fim-rate2} shows all models irrespective of FIM rate have a similar performance.

On the other hand, we find that the FIM rate does significantly affect infilling capabilities. Even though the gain in FIM perplexity in Figure \ref{fig:pretraining:fim-rate1} due to a higher FIM rate is negligible, increasing this rate yields a consistent improvement in the infilling pass rate as shown in the right plot in Figure \ref{fig:pretraining:fim-rate2}. This indicates that to investigate the FIM capabilities of our models, it is not sufficient to consider language modelling perplexity measures such as test loss, but we should also consider non-loss based evaluations.

In Appendix \ref{appendix:more_fim_rate}, we show further evidence across a range of scales that higher FIM rates improve infilling performance but that this gain is not reflected in the perplexity evaluation. 

Given the results here and in Appendix \ref{appendix:more_fim_rate}, it is natural to question why we train our core series of models with a FIM rate of 50\% rather than 90\% or higher. Models with a FIM rate of 90\% show superior performance while maintaining the FIM-for-free property. This was mainly accidental, as we had already trained the main series prior to seeing the FIM rate ablation results,\footnote{In particular, our earlier ablations based only on loss had indicated that the gains from increasing the FIM rate to 90\% should be negligible, resulting in us choosing a more moderate value of 50\%. More detailed study using all 3 infilling benchmarks showed that there is in fact a noticeable gain in using even a higher FIM rate.} and it was prohibitively costly to retrain all the models with the higher rate.  

The results here motivated us to train a second 6.9B FIM model with a FIM rate of 90\% on code to obtain the strongest infilling model to date at this scale. The comparison of results is found in Table \ref{tab:top-models}. We note however from Figure \ref{fig:fim-rate:pass-rate} that a FIM rate of 50\% does not seem to be too far from optimal. 

\begin{figure}[ht!]
\centering
\includegraphics[width=\textwidth]{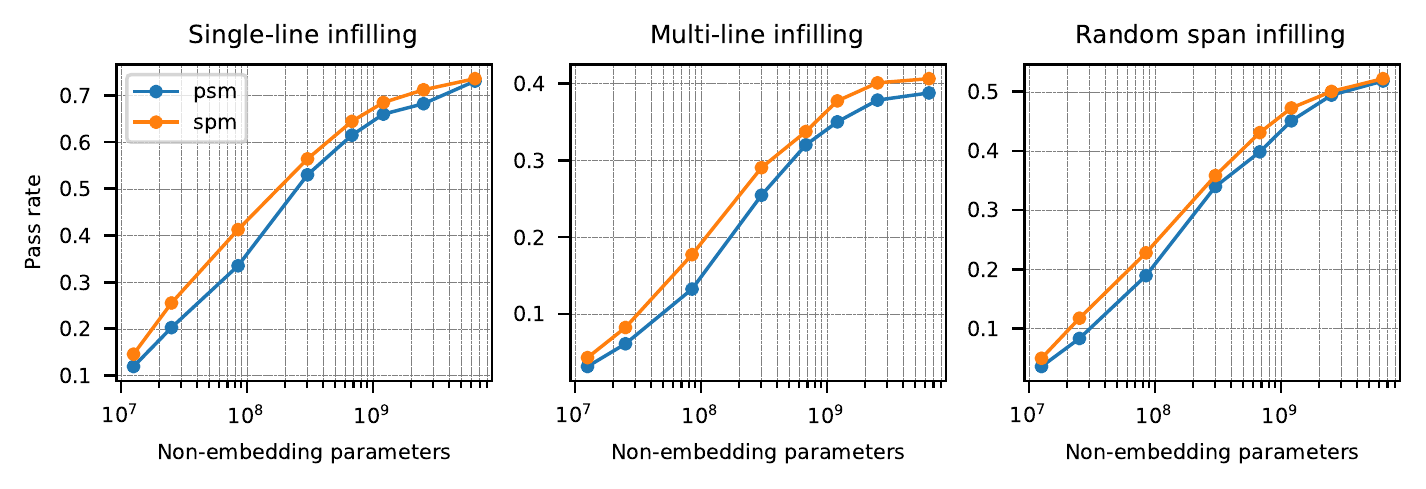}
\caption{SPM mode shows a slight advantage in performance across scale. All the evaluations in this plot are at temperature 0.2 and 100 samples per task for single-line and multi-line infilling and 200 samples per task for random span infilling.}
\label{fig:pretraining:psm-spm}
\end{figure}
\subsection{SPM vs PSM vs joint SPM+PSM training}
\label{sec:pretraining:psm-spm}
In Section \ref{sec:fim_training}, we describe two ways of constructing a FIM example: 
$[\text{suffix}, \text{prefix}, \text{middle}]$ and
$[\text{prefix}, \text{suffix}, \text{middle}]$.
Here we study how this choice affects performance  during pretraining and evaluation.

The main finding is that SPM is slightly stronger than PSM in our benchmarks in general as evidenced by Figure \ref{fig:pretraining:psm-spm}. We train a series of FIM models with a FIM rate of 50\% with the FIM rate equally allocated to PSM and SPM. We find that evaluating these models in SPM mode yields a consistently higher performance than PSM across scale. This is likely due to the fact that in SPM, there is no distinction between the prefix and the middle sections as they are one contiguous sequence of text. This makes it more natural for the model to continue from the prefix in contrast to PSM where attention has to first identify where the span token is.

However, this does not imply that we should train solely on SPM. In Table \ref{tab:pretraining:psm-spm}, we train large models on pure PSM, pure SPM, and our default 50-50 SPM+PSM mix, and evaluate them in all modes. We observe a positive transfer of capability between PSM and SPM. Training joint FIM with a 50\% FIM rate obtains roughly the same performance in SPM mode as training pure SPM FIM with a 90\% FIM rate. Not only is joint pretraining the most efficient, but it also yields the most flexible model with two inference modes.

It is noteworthy that the recent infilling works using data transformations similar to FIM such as \citep{donahue, cm3, incoder} utilize a format similar to PSM. The above findings indicate that this choice leads to suboptimal infilling performance.

\begin{table}[ht!]
\centering

\begin{tabular}{cccccccc}
\hline\Tstrut
Train distribution & FIM rate &  \multicolumn{2}{c}{Single-line} &\multicolumn{2}{c}{Multi-line } & \multicolumn{2}{c}{Random span } \\ 
 & & PSM & SPM & PSM & SPM & PSM & SPM \\[0.1cm]
\hline \\ [-0.2cm]
Joint & 0.5 & 0.550 & 0.595 & 0.265 & 0.293 & 0.367 & 0.379 \\
Joint & 0.9 & 0.616 & 0.622 & 0.290 & 0.305 & 0.397 & 0.420 \\
PSM & 0.9 & 0.583 & 0.625 & 0.273 & 0.305 & 0.362 & 0.274 \\
SPM & 0.9 & 0.023 & 0.586 & 0.008 & 0.301 &  0.007 & 0.386 \\
\hline
\end{tabular}
\caption{Comparison of FIM performance when trained and evaluated in various SPM, SPM settings.  All the joint runs put 50\% of the total FIM rate on PSM and 50\% on SPM. All results are obtained with temperature 0.2 and 100 samples per task.  }
\label{tab:pretraining:psm-spm}
\end{table}

\input{doc-context-pass-rate-figure}

\input{doc-context-loss-figure}

\subsection{Context-level vs document-level FIM}
\label{sec:pretraining:context}

In Section \ref{sec:fim_training}, we noted two ways of implementing FIM, context-level and document-level FIM, where augmentation is applied either before or after packing and chunking. We now ablate this choice on a series of code models trained with a 50\% FIM rate and the default joint PSM-SPM mix.

In Figure \ref{fig:pretraining:doc-context}, we find that context-level FIM yields a consistent and significant improvement over document-level FIM across all the range of scale. This is a noteworthy contrast to the perplexity evaluation in Figure \ref{fig:pretraining:doc-context-loss} (right) where the improvement is an almost negligible 0.001 nats/token. This corroborates the finding in Section \ref{sec:pretraining:fim-rate} that perplexity evaluation does not always capture the gains in the sampling performance.

Also, we previously explained that document-level FIM can result in fragmented FIM data with a missing prefix and/or suffix from the chunking step of data loading pipeline. Figure \ref{fig:pretraining:doc-context-loss} (left) shows that training on these invalid examples in document-level FIM does not affect the left-to-right evaluation. Hence, practitioners might  still sometimes prefer document-level FIM due to its simpler implementation.

\subsection{Middle span selection}
\label{sec:pretraining:span}

An important consideration in FIM training is the choice of middle span. In this work, the middle span is chosen uniformly at random where the split between prefix, middle, suffix happens at the character level. In this section, we examine this choice. Instead of trying FIM across syntactic boundaries, such as functions and class bodies, we restrict our ablations to simple, generalizable approaches which are language agnostic. We select spans in three different ways, splitting randomly by lines, tokens, and characters. The section boundaries are selected uniformly at random from the allowed splitting positions based on the span type. Here, a token refers to a word in the byte-pair encoding (BPE) vocabulary. In practice, this is implemented by applying the FIM augmentation after the documents are encoded with BPE (see Appendix \ref{appendix:impl_details}). For simplicity, we run all our experiments in PSM mode in this ablation.

In Table \ref{tab:pretraining:span-selection} we see that training only on the line-based middle spans gives the models a slight advantage in the single-line and multi-line infilling benchmarks. This is not surprising since these evaluations are completely in distribution with line based middle span runs. On the other hand, the line based training fails almost completely in the random span infilling benchmark. Interestingly, the advantage provided in line-based evaluations from concentrating all the FIM distribution on line based middle spans in training is quite small relative to how much it hurts the model in random span infilling benchmark.

Training with token-level random spans does slightly better on random span infilling, but is still not competitive compared to character-level runs on this benchmark. The reason is that token-level FIM models  are never trained on cases where a token is broken into two parts across the boundaries of middle with prefix or suffix. When the middle section is selected completely at random at the character level, subtokens are introduced naturally at the beginning and the end boundaries of the middle section. There is no train-test mismatch and the model is able to understand and solve more random span infilling tasks while still performing well in single-line and multi-line infilling.

\begin{table}[ht!]
\centering
\begin{tabular}{cccc}
\hline\Tstrut
Training middle span & Single-line infilling & Multi-line infilling & Random span infilling \\[0.1cm]
\hline\Tstrut
Line-level random span & 0.586 & 0.269 & 0.015 \\
Token-level random span & 0.548 & 0.242 & 0.102 \\
Character-level random span & 0.557 & 0.250 & 0.321 \\
\hline
\end{tabular}
\caption{Pass rates of medium models pretrained with various middle span selection strategies. Training on line-based spans improves the single- and multi-line infilling metrics reported in InCoder, but line- and token-level spans used in previous works can not robustly handle real life use cases where the span starts or ends in subtokens. Overall, character-level random span run dominates in random span benchmark while it is also not far behind in single and multi line infilling.}
\label{tab:pretraining:span-selection}
\end{table}

\section{Finetuning results}
\label{sec:finetuning}

\begin{figure}[ht!]
\centering
    \begin{subfigure}[b]{\textwidth}
        \centering
        \includegraphics[width=\textwidth]{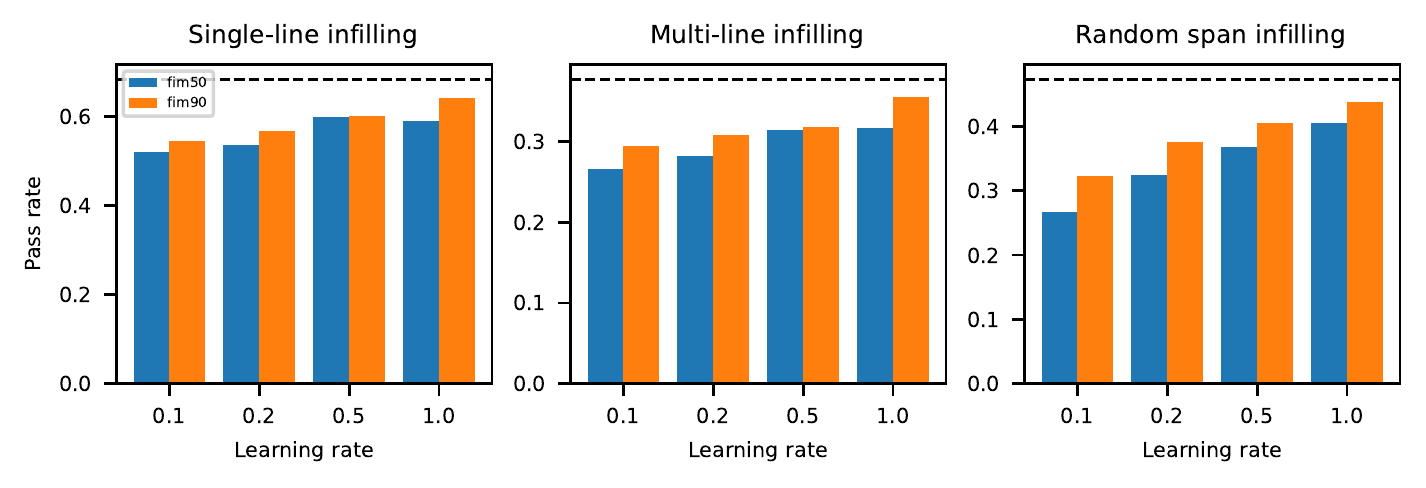}
        \caption{25B tokens of FIM finetuning.}
    \end{subfigure}
    \begin{subfigure}[b]{\textwidth}
        \centering
        \includegraphics[width=\textwidth]{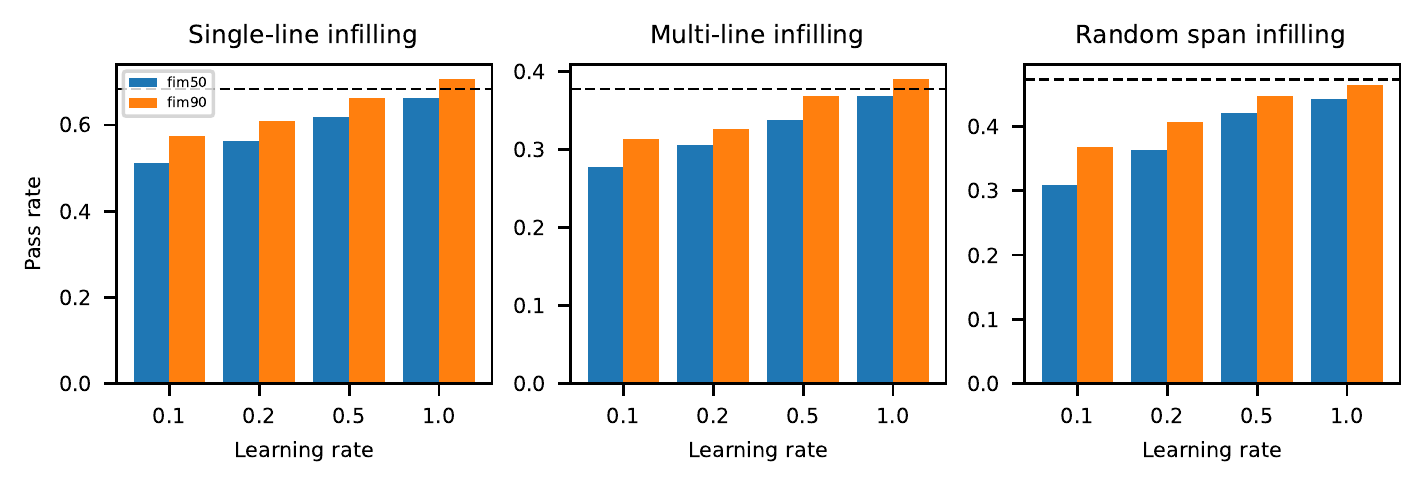}
        \caption{50B tokens of FIM finetuning.}
    \end{subfigure}
\caption{Evaluation of the final snapshots of models pretrained for 100B tokens without FIM and then finetuned for 25B (row a) and 50B (row b) tokens with FIM. The x-axis shows the learning rate multiplier relative to the pretraining learning rate. The dashed line indicates the baseline performance of the model pretrained for 100B tokens with a FIM rate of 50\% with no additional finetuning. Only the most aggressive combination of 90\% FIM rate and a learning rate multiplier of 1.0 with 50B tokens of finetuning catches up to the performance of the baseline. Reported results are with temperature 0.2 and 100 sampler per task.} \label{fig:finetuning_main_fig}
\end{figure}
In this section, we investigate whether we can finetune existing AR models to learn the FIM capability. Ideally, after finetuning, an AR model would reach the same level of performance on FIM evaluations as it would have achieved if it were pretrained with FIM. Given that FIM can be learned during pretraining without extra compute cost, it is natural to expect that the model should also be able to learn this task quickly in finetuning. Surprisingly, we find that for finetuned models to reach the same level of performance as baseline pretrained models, one needs to expend a large amount of compute relative to the pretraining compute.

To show this, we finetune an XL model pretrained for 100B tokens without FIM using different choices of finetuning hyperparameters. Specifically, we train 16 finetuned models with 4 choices of learning rates (0.1, 0.2, 0.5, 1x multiples of pretraining learning rates), 2 different FIM rates (0.5 and 0.9), and 2 different choices of finetuning horizons (25B and 50B tokens). We use this large variety of hyperparameter choices to both ensure that our conclusion is robust and to better understand the effect of hyperparameters on the final performance. The results are summarized in Figure \ref{fig:finetuning_main_fig} where we compare the performance of these 16 models with that of the XL model trained for 100B tokens with a FIM rate of 50\% without any finetuning. It is evident from this figure that
even with significant additional finetuning compute, AR models finetuned with FIM do not reach the same performance as the models pretrained with FIM (and without any FIM finetuning).

Among these 16 models, the only setting where the gap between pretrained baseline and finetuned models is closed is the 50B token run with a FIM rate of 0.9 and learning rate multiplier of 1.0 relative to pretraining. More generally, we find that higher learning rate, FIM rate, and longer finetuning all seem helpful for improving FIM performance in finetuning.

We find it particularly surprising that such high learning rates and lengthy finetuning are necessary for reaching the similar level performance. We discuss this topic more in Section \ref{sec:discussion}. We note that although reaching the same level of performance as in pretraining requires a large amount of compute, a small amount of finetuning (especially with high FIM and learning rate) is still sufficient for the model to reach non-trivial levels of FIM performance on our metrics. We present further results on dynamics of finetuning in Appendix \ref{appendix:ft_dynamic}.

\section{Discussion}\label{sec:discussion}
 
 \textbf{Pretraining vs finetuning.} In the previous sections, we studied how to efficiently teach FIM to causal language models. A main finding was that FIM can be learned for free in  pretraining. In contrast,  we saw in Section \ref{sec:finetuning} that learning FIM in finetuning can be quite expensive. Here we describe some potential explanations for these findings. 
 
 The main intuition for why FIM can be learned for free in pretraining is that breaking a document into three pieces and shifting the middle one to the end effectively creates three smaller documents. In particular, each piece still requires predicting next tokens from left to right, keeping the total number of tokens processed autoregressively the same.

\begin{figure}[ht!]
\centering
\includegraphics[width=0.85\textwidth]{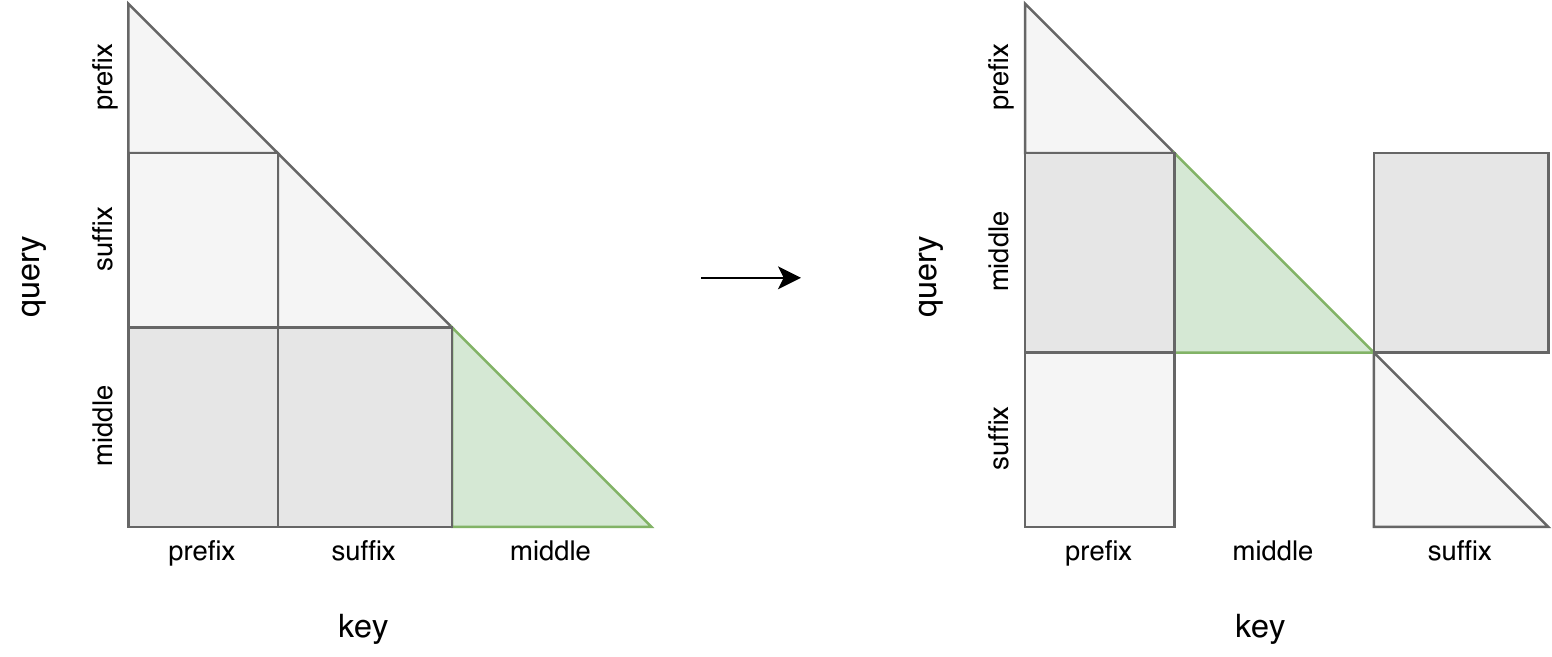}
\caption{Visualization of causal attention pattern of FIM data. Unraveling both the query and key embeddings back in the canonical left-to-right order shows that FIM allows the transformer to attend to future context when decoding the middle section without complex architectural changes. One side-effect is that the suffix probability no longer depends on the middle span.
}
\label{fig:fim-attn}
\end{figure}

On the other hand, even though FIM data is locally identical to autoregressive data, FIM does impose a different global attention pattern over the whole document. To visualize this, we show the causal attention mask of a FIM document in Figure \ref{fig:fim-attn}. These new attention pattern could be the reason why it takes a relatively long token horizon and a high learning rate to learn FIM in finetuning. It is possible that there is ossification \citep{hernandez2021} in the learned document-wide attention pattern in regular AR pretraining which requires a lengthy finetuning stage to adapt to the attention pattern needed in FIM.
 
\textbf{FIM loss, AR loss, and the difficulty of FIM task.} Naively, since FIM does not come at a cost in AR capability, one may expect FIM to be an easy task.  In fact, the opposite seems to be the case. There is substantial evidence that FIM can often be much harder than normal left-to-right generation.

Intuitively, it is often easier to continue a text in a plausible manner than to continue the text conditioned on ending in a specific suffix. The latter requires planning a plausible narrative connecting the two pieces, starting the generation in a way that matches the prefix, and \emph{stopping the generation at the right time} so it connects to the suffix. In particular, in FIM the model is trained to generate $\textsc{<eot>}$ when the middle ends and connects to the suffix. On the other hand, when the model fails to produce $\textsc{<eot>}$ in the allotted budget, we often end up with truncated samples which do not connect well to the suffix. For example, consider the following:

\vspace{-6pt}
\begin{center}
\begin{minipage}{5in}
\begin{lstlisting}[style=text]
When I was young, I only liked to pl`ay video games. Over time, I started thinking if it'd be possible to make bots to play better than any human can ever play these games. I eventually decided I liked working on the latter more than playing the games themselves` and that's how first I got interested in AI research.

When I was young, I only liked to pl`ay video games. I would play sometimes more than 13 hours per day. The rush, novelty, and variety were beyond anything real life could offer. I loved the challenge and I excelled at it. I would often skip classes and go to` and that's how first I got interested in AI research.
\end{lstlisting}
\end{minipage}
\end{center}

Both completions above connect well to the prefix, but only the first manages to connect well to the suffix. The second completion in contrast fails to produce $\textsc{<eot>}$ in the allotted budget resulting in a bad sample.\footnote{Even though the completion may have been able to connect to the suffix with a bigger budget, the challenge is it is unclear how much budget is enough. In practice, often a reasonable budget for the maximum number of tokens for the middle must be imposed.} This turns out to be a common failure in FIM sampling. Even though, left-to-right sampling also struggles sometimes with related issues, this type of failure is more troublesome in FIM since a failure to connect to the suffix cannot easily be fixed by post-processing. For example, trimming the sample to the last paragraph or line is often an effective way in improving sample quality in AR sampling, but does not help in FIM. We discuss this and other issues associated with FIM sampling more extensively in Appendix \ref{appendix:qual_eval}. 

\begin{figure}[ht!]
\centering
\includegraphics[width=\textwidth]{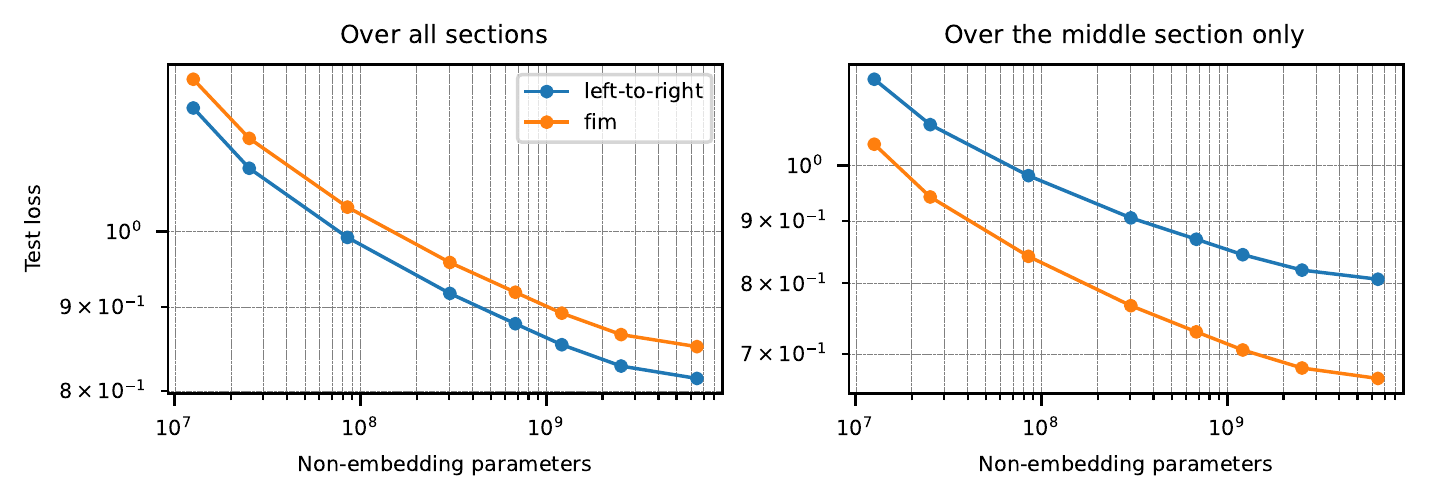}
\caption{Comparison of the overall (left) and middle span (right) loss of 50\% FIM code models.
In the left plot, we see that the AR loss is consistently lower than the FIM loss suggesting that next-token prediction is inherently more compressible than infilling in the middle.
The right figure  evaluates the conditional loss of the middle span given the surrounding context showing that ~$P_{\text{FIM}}(\text{middle} |  \text{prefix}, \text{suffix}) >  P_{\text{AR}}(\text{middle} |  \text{prefix})$. Here, FIM attains a lower loss because it can attend to the suffix. We emphasize that  left-to-right and FIM here do not refer to model type, as all models in this figure are FIM models. They refer rather to the type of test loss used in evaluation.
}
\label{fig:fim-ar-loss}
\end{figure}

The difficulty of FIM task compared to AR task is also reflected in the loss associated with each task. To see this, in Figure \ref{fig:fim-ar-loss}, we compare the FIM loss with the AR loss over a suite of FIM models all with 50\% FIM rate. To remove confounders, we ensure the documents that underlie the AR test set are the same documents that are transformed through FIM to make up the FIM test set. We find the FIM perplexity is consistently higher than the AR perplexity across scale.  That is, on average
\[ P_{\text{FIM}}( [\text{prefix}, \text{suffix}, \text{middle}]) < P_{\text{AR}}([\text{prefix}, \text{middle}, \text{suffix}]), \]
which means the models have a harder time modelling the same document in FIM format than AR format. 

\textbf{Context-level vs document-level FIM and FIM rate.} In Section \ref{sec:pretraining:context}, we saw that context-level FIM typically outperforms document-level FIM. Here, we note a connection between this finding and the results in Section \ref{sec:pretraining:fim-rate} and Appendix \ref{appendix:more_fim_rate} about FIM rate.

The basic observation is that document-level FIM effectively leads to a \emph{lower FIM rate} compared to context-level FIM, even with the same nominal value of FIM rate. As a thought experiment, consider the setting where all the documents in the training dataset are much longer than the context size. In this setting, when using document-level FIM, the model almost never sees the prefix, middle, and suffix of the same document appear in the same context together after chunking. As such, we would expect the model to struggle to learn FIM in this setting. In less extreme situations,  there are many documents  shorter than the context size and hence the above phenomenon is less pronounced. Still, because of long documents in training data and the usual artifacts of document packing, document-level FIM results in a lower effective FIM rate. Here, we define the effective FIM rate as the fraction of examples that are in FIM format and with all three of the prefix, middle, and suffix appearing within the same context. 

This decrease in effective FIM rate is likely the main reason behind the stronger performance of context-level FIM in Section \ref{sec:pretraining:context}. We note that the exact amount of decrease in effective FIM rate depends on the details of distribution of document lengths. It is important to remember that even if the data distribution does not have many long examples, the decrease in effective FIM rate will still be present because of document packing.

\section{Related work}\label{sec:related_work}

Masked language modeling is closely related to text infilling in that consecutive runs of masked tokens can be interpreted as spans that the model must infill. While early masked language models like BERT \citep{bert} masked tokens randomly, T5 \citep{T5}, SpanBERT \citep{spanbert}, and BART \citep{bart} demonstrated improvements when contiguous runs of tokens are masked. However, because these models focus on representation learning, the span lengths are typically much shorter than a sentence or even a single line of code. Within our modalities of interest, DOBF \citep{dobf} trains BERT on code, and HTLM \citep{htlm} trains BART on HTML data.

Text infilling can also be seen as a special case of autoregressive language modeling where the standard left to right generation order is replaced by a more flexible ordering. XLNet \citep{xlnet} modifies the attention mask in a standard transformer to enable token generation in any user-specified order, while Insertion Transformer \citep{insertion-transformer}, KERMIT \citep{kermit}, and InDIGO \citep{indigo} allow the model to predict a location for the next token before predicting the token. Similarly, Blank Language models \citep{blanklm} generate text by iteratively selecting a blank and replacing it with a token (and optionally more blanks).

Similar to our work, \cite{zhu-infilling}, \cite{donahue}, GLM \citep{glm}, CM3 \citep{cm3}, and InCoder \citep{incoder} utilize left-to-right autoregressive modeling by moving the infill regions to the end of context, with regions separated by sentinels. Notably, \cite{donahue} explore infilling spans of varying granularities, such as words, sentences, or paragraphs, and InCoder \citep{incoder} uses a similar evaluation framework to ours by studying infilling capabilities on sampling based benchmarks created from HumanEval \citep{codex}. While several of these works support infilling multiple spans, we focus on the single span setting for practicality (e.g. in computer-based text generation, where the placement of cursor implies the location we want to infill). Additionally, our paper emphasizes the computational efficiency of training for infilling at scale. While we do not study syntactically or semantically motivated infilling spans, we show selecting spans at the character level improves the robustness of infilling.

Text infilling can also be performed using a GAN \citep{maskgan}, but REINFORCE is required to deal with the discreteness of text. Text infilling can also be done through gradient search \citep{tigs}, where tokens within the infilled span are optimized with gradient descent and collapsed to the nearest neighbor.

Overall, there are two approaches for imbuing models with infilling capabilities: first, through new architectures like SpanBERT and XLNet; second, through data formatting. In general, the latter approach can be seen as altering the behavior of a language model through control codes, which was motivated in CTRL \citep{ctrl} to improve the steerability of generation. DistAug \citep{distaug} is another related work that trains jointly on transformed data while conditioning on the transformation type. While infilling is a specific use case that can be realized through both architecture and data, it is generally easier and more universal to learn additional skills by introducing new training distributions than hardwiring them.

The strongest infilling system at scale to our knowledge currently is  code-davinci-002 released this past March \citep{edit_insert}. The present paper describes some of the early research that went into powering the infilling capabilities of this more powerful model. In Appendix \ref{tab:top-models}, we present a comparison between this system, our 6.9B models, and the InCoder 6.7B model on our infilling benchmarks. 

\section{Conclusion}\label{sec:conclusion}

In this work, we show that causal decoder-based language models can learn to fill in the middle of a document after being jointly trained on a mixture of traditional left-to-right  and FIM transformed data. A single FIM model can import modules, write docstrings, and complete functions, subsuming specialized models finetuned for individual tasks \citep{codex}, providing substantial extra capability over traditional left-to-right language models.

One important finding here is the FIM-for-free property. Figures \ref{fig:no-harm-perp} and \ref{fig:fim_loss} show that with the same amount of compute, FIM models achieve the same test loss as AR models on left-to-right test loss while achieving lower FIM loss. This is further strengthened using non-loss based evaluations in Section \ref{sec:pretraining}.

We also investigate FIM finetuning since a lot of the existing language models do not have FIM capabilities. Our results demonstrate that a canonically pretrained left-to-right model does not acquire the new skill to the fullest extent of the given model size even with careful hyperparameter tuning and a significant amount of finetuning compute relative to pretraining. This suggests that for the best FIM performance, pretraining jointly from scratch with our recommended hyperparameters is more effective than finetuning.

To study FIM capabilities precisely, we use the infilling code benchmarks from InCoder \citep{incoder} and introduce the new random span infilling benchmarks based on HumanEval \citep{codex}. From these, we learn a few important lessons.
First, perplexity does not reflect the true infilling performance, and one should design the infilling benchmarks carefully to measure progress. Second, FIM capabilities depend considerably on the FIM rate and implementation like context-level FIM but left-to-right capabilities are unaffected by these choices as long as the FIM rate is kept below 100\%. Third, applying FIM at the character level imbues the model with natural robustness to subtokens and makes it possible to deploy the model in the wild, for example, as a coding assistant. 

All in all, we show FIM models are strictly more capable than canonically trained left-to-right models, at least within the bounds of the evaluations we consider, and we demonstrate how to train FIM models efficiently and competitively.

\subsection{Recommended FIM hyperparameters}
\label{sec:conclusion:recommendation}

In Section \ref{sec:pretraining}, we see there are a number of hyperparameters in training FIM models. In all cases, we recommend applying FIM transformation at the character level and always including some character-level random spans as it allows the model to generate sensible completion even when the prefix and suffix end in the middle of a token. We note that for mid-token robustness, inference in PSM mode can be superior to the particular SPM mode explored in this work. However, pretraining with joint PSM and SPM yields the best performance due to a positive transfer between the two formats. In terms of implementation, context-level FIM is superior but document-level FIM is also an option if a simpler implementation is desired. Finally, we observe improved performance even up to a FIM rate of 90\% without any cost in AR capabilities. In practice, any value in the range between 50\% and 90\% is a reasonable choice. Note that this is in contrast with some related prior work such as \citep{incoder} which typically uses lower values of FIM rate such as 15\%, which our results indicate to be suboptimal.

\subsection{Future directions}
\label{sec:conclusion:future}

There are several important related directions that we did not cover here. For example,

\begin{enumerate}
\item \textbf{Smarter span selection}: 
We only consider spans selected uniformly at random for generality, but mixing in semantically or syntactically meaningful spans \citep{donahue,spanbert,reasonbert} can considerably improve infilling performance. In Section \ref{sec:pretraining:span}, we see that training on line-level spans instead of character-level spans improves line-based infilling results. In our preliminary experiment, selecting the middle span to be exactly one word was shown to significantly improve accuracy on cloze-like tasks.
Smarter span selection involves language specific parsing and new benchmarks which may be tricky to make, but we expect this to produce stronger FIM models.
\item \textbf{Steerable generation}:
FIM models generate spurious content or struggle to generate a sensible completion in the allotted token budget because they do not know the length or the style of infilling the user desires. Applying ideas like RL from human feedback \citep{rlhf} and instruction following \citep{if}, among other methods of controllable generation, could address this issue by providing further alignment with the users' intent.
\item \textbf{Further examination of the FIM-for-free property}: Even though we provide substantial evidence for the FIM-for-free property, we cannot completely rule out that there are  benchmarks not considered here where FIM models underperform AR models. As such, further strengthening or refuting the FIM-for-free property remains an interesting direction. 
\item \textbf{Multiple infilling slots}:
Many prior works in infilling explored multiple infilling slots \citep{T5,incoder}. We do not study this, as there are already a number of considerations in training single-slot models, and inference challenges unique to infilling. Furthermore, in most applications, we anticipate single-slot infilling to be just as useful. We anticipate the inference challenges and failure modes to increase when considering multi-slot infilling. To make progress in multi-slot infilling however, creating appropriate sampling-based benchmarks is essential, as perplexity based evaluation would be increasingly unhelpful.  There is a vast design space for these benchmarks and a vast array of extra training hyperparameters when going from single-slot to multi-slot infilling. 

\item \textbf{Improving natural language FIM performance}:
Qualitatively, our FIM models tend to perform better in code than language. This is perhaps not surprising given that code is a formal language, and as such, has more structure and less uncertainty. Improving infilling performance on natural language is an interesting future direction, but can be tricky because evaluation of free-form generation in language is not as straightforward as measuring functional correctness in code. We expect training on more semantically meaningful or shorter spans can help here but it is unclear what test distribution to use and how to evaluate this well in the general case.

\item \textbf{Role of bidirectionality and attention}:
There is much to be understood in the role of attention and the training objective in free-form infilling performance. In this work, we use decoder based language models, which are currently the dominant paradigm of large scale language modelling. However, it is possible that from the point of view of infilling, other training objectives and architectures are superior. In this direction, \citep{bidirlm} show a BERT style architecture performs better than FIM-like models but the results are mostly limited to single-token infilling. A more systematic study, similar to \citep{huggingface, unifying} but focused on free-form infilling generation, can clarify this further.
Somewhat related to this, it is interesting to investigate the interaction of absolute and relative positional embedding and their variants with FIM. Preliminary results, not reported here, indicate that the FIM-for-free property still holds with absolute positional embedding. 

\end{enumerate}

Finally, our experience with the FIM-for-free property brings up the intriguing question of \emph{what other useful skills can be learned jointly with no or little cost to the original capabilities of language models}. There have been a number of interesting works on this topic and we anticipate even more to follow, but many works often omit critical analysis for more broad adoption and comparison. We propose the following methodology to help advance research toward answering this question:
\begin{enumerate}
\item Establish a \emph{budget} in the amount of original capabilities that one is willing to sacrifice to learn a new capability.
\item Maximize the new capability within this budget.
\end{enumerate}
The budget-capability trade-off is not only theoretically interesting but also practical, allowing researchers to integrate new capabilities based on proper trade-off analysis. We look forward to a future where large language models have increasingly diverse and high value capabilities.

\section*{Acknowledgments}

We would like to thank Shantanu Jain, Alex Paino, Alec Radford, Nick Ryder, Pranav Shyam, and Qiming Yuan for useful discussions and help at various stages of the project. We are also grateful to Christina Kim, Rachel Lim, Andrew Mayne, Maddie Siemens, and Natalie Staudacher for the help with the API infrastructure and qualitative evaluation of FIM, and to Angela Jiang, Katie Mayer, Rajeev Nayak, Henrique Pond\'e, and Felipe Such for invaluable work and immense effort on deployment. Finally, we thank Bob McGrew and Wojciech Zaremba for unceasing support throughout the project, and Karl Cobbe, Angela Jiang, Alec Radford, and Pranav Shyam for their valuable feedback on the paper.

\bibliographystyle{abbrvnat}
\bibliography{bibliography}

\appendix

\clearpage

\section{Architecture and datasets}\label{appendix:arch}
\label{sec:setup:model}
We use 8 causal transformer decoder models \citep{transformer} with similar architecture, optimization hyperparameters, and encoding to Codex and GPT-3 \citep{codex, gpt3}. The main architectural details of our models are summarized in Table \ref{tab:arch}. The only architectural modification we introduce is the use of relative attention \citep{relattn,xl} rather than learned positional embeddings. This increases the parameter count negligibly but leads to improved performance. We also increase the learning rates of our three largest models by a factor of 2 for improved final performance, as it is known that GPT-3 series of models use rather conservative choices of learning rates. The context size for all the models is $2048$.

We train our code models on the same dataset that was used to train Codex, which is a 159 GB Python dataset scraped in May 2020. As such, we expect no train set contamination from the subsequent public release of HumanEval. Similar to GPT-3 and unlike Codex, we train our models from scratch from a random initialization. All models from the main scans are trained for 100B tokens irrespective of size. Due to this fixed token budget, we expect our largest models to be undertrained \citep {chinchi} and to benefit significantly from longer training. For our natural language models, we use the same dataset as was used in GPT-3 \citep{gpt3}, the details of which are described in Section 2.2 of that paper.

\begin{table}[ht!]
\centering
\begin{tabular}{ccccccccc}
\hline\Tstrut
Model Name  & $n_{\text{param}}$ & $n_{\text{ne}}$ & $n_{\text{layers}}$ & $d_{\text{model}}$ & $n_{\text{heads}}$ & $d_{\text{head}}$ & Batch Size &  Learning Rate  \\[0.1cm]
\hline\Tstrut
XXS & 50M & 11M &  6 & 384 & 6 & 64 & 0.5M & $1.6 \times 10^{-3}$  \\
XS & 77M & 26M & 8 & 512 & 8 & 64 & 0.5M & $1.4 \times 10^{-3}$  \\
Small & 164M & 87M & 12 & 768 & 12 & 64 & 0.5M & $6.0 \times 10^{-4}$  \\
Medium & 411M & 308M & 24 & 1024 & 16 & 64 & 0.5M & $3.0 \times 10^{-4}$  \\
Large & 844M & 689M & 24 & 1536 & 16 & 96 & 0.5M & $2.5 \times 10^{-4}$  \\
XL & 1.4B & 1.2B & 24 & 2048 & 16 & 128 & 1M & $4.0 \times 10^{-4}$  \\
2.8B & 2.8B & 2.6B & 32 &  2560 & 32 & 80 & 1M & $3.2 \times 10^{-4}$  \\
6.9B & 6.9B & 6.5B & 32 &  4096 & 32 & 128 & 2M & $2.4 \times 10^{-4}$  \\
\hline
\end{tabular}
\caption{The model architecture for our suite of models. The 6 largest models follow similar architecture as models Small to 6.7B in the GPT-3 paper. The differences in the tables are due to minor calculation errors and typos in Table 2.1 of that paper. The $n_{\text{param}}$ column has the total number parameters in each model while $n_{\text{ne}}$ column has the number of parameters excluding the embedding and unembedding layers. Following \citep{scaling_laws}, we use the number of non-embedding parameters in our scaling plots. We do not tie the weights in the embedding and unembedding layers. } 
\label{tab:arch}
\end{table}

\section{Scaling trends for FIM rate ablations}\label{appendix:more_fim_rate}
\begin{figure}[ht!]
\centering
\includegraphics[width=\textwidth]{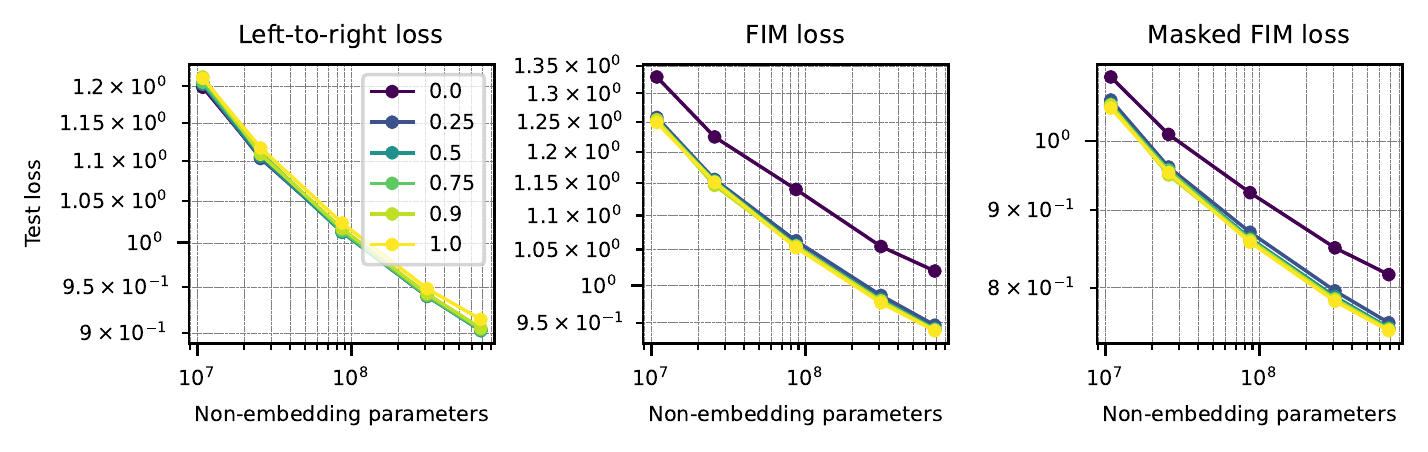}
\caption{Comparison of model scaling trends of perplexity with varying FIM rates. Left-to-right loss does not have a noticeable degradation unless a FIM rate of 100\% is used (left). We also find that the FIM losses are similar to one another when the model is trained with some FIM transformations (middle and right).}
\label{fig:fim-rate:loss}
\end{figure}

\begin{figure}[ht!]
\centering
\includegraphics[width=\textwidth]{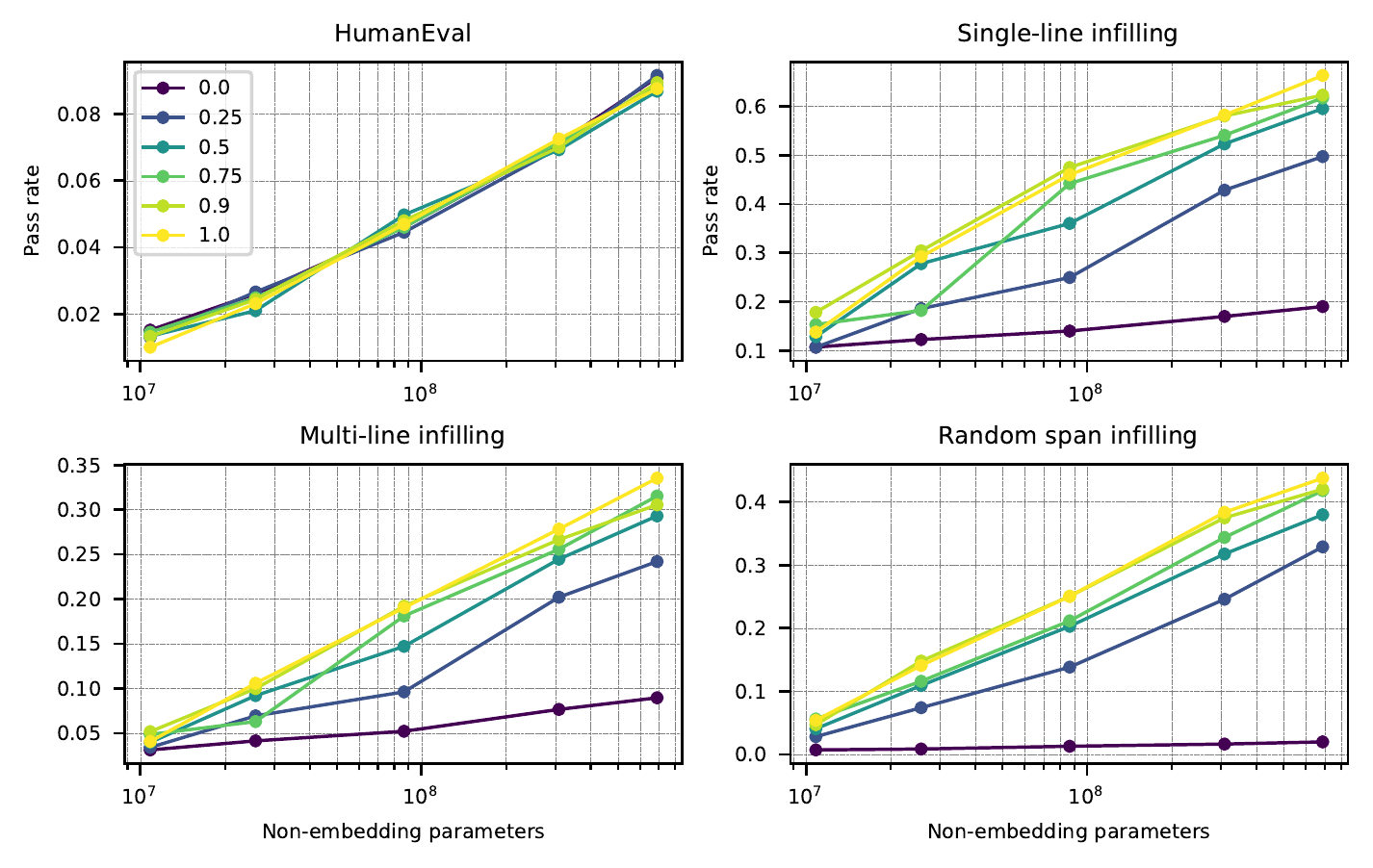}
\caption{Comparison of model scaling trends of sampling evaluation with varying FIM rates. While increasing the FIM rate has no effects on HumanEval, it does result in consistent gains on the infilling benchmarks with no noticeable improvement after 90\% FIM. 
At a first glance, it may seem counterintuitive that left-to-right models can solve a nontrivial number of problems in single- and multi-line benchmarks. This is not a bug, but a feature. We sample in SPM mode and some line-based infilling problems have empty or extraneous suffixes.
To obtain these results, HumanEval was evaluated with temperature 0.8 and 500 samples per task to reduce variance. All infilling benchmarks, having much more problems than HumanEval, were evaluated with temperature 0.2 and 200 samples per task.
}
\label{fig:fim-rate:pass-rate}
\end{figure}

In Section \ref{sec:pretraining:fim-rate}, we see higher FIM rate improving the FIM performance of our models without impacting the original capabilities. This conclusion was based on the learning curves of HumanEval and light random span infilling pass rates measured with a small number of samples during pretraining. To further demonstrate this claim, we train a series of models for 50B tokens with FIM rates: 0, 0.25, 0.5, 0.75, 0.9, and 1.0. In Figure \ref{fig:fim-rate:loss} and \ref{fig:fim-rate:pass-rate}, we present the model scaling trends of perplexity and sampling evaluation when different FIM rates are used.

Again, we find that transforming a high fraction of training data into FIM does not result in a degradation in the original capabilities as measured by the test loss and HumanEval pass rate. The only noticeable degradation is observed in perplexity evaluation at 100\% FIM rate. As for FIM capabilities, increasing the FIM rate yields a significant improvement on the infilling benchmarks and can change the slope of model scaling trends of pass rates. However, a high FIM rate does not lead to a commensurate reduction in FIM losses, which corroborates that perplexities do not always capture real world performance.

\section{Details of FIM implementation}\label{appendix:impl_details}

When FIM is applied at the document level before packing, both character-level and token-level FIM is straightforward to implement. We simply choose two positions at random to break a document into three sections, and format them as a FIM document. Only the order of encoding and splitting changes as shown in the python pseudocode below:

\vspace{-6pt}
\begin{center}
\begin{minipage}{5in}
\begin{lstlisting}[style=python]
def token_level_psm_fim(document: str, vocab: Vocab) -> List[int]:
    tokens = vocab.encode(document)
    prefix, middle, suffix = randomly_split(tokens)
    return [
        vocab.sentinel("prefix"), *prefix,
        vocab.sentinel("suffix"), *suffix,
        vocab.sentinel("middle"), *middle,
    ]

def character_level_psm_fim(document: str, vocab: Vocab) -> List[int]:
    prefix, middle, suffix = randomly_split(document)
    return [
        vocab.sentinel("prefix"), *vocab.encode(prefix),
        vocab.sentinel("suffix"), *vocab.encode(suffix),
        vocab.sentinel("middle"), *vocab.encode(middle),
    ]
\end{lstlisting}
\end{minipage}
\end{center}
\vspace{-3pt}

In contrast, applying the transformation after packing and chunking as in context-level FIM can be somewhat tricky depending on the choice of middle span. As previously mentioned in Section \ref{sec:fim_training}, the input context to the model is first split around the $\textsc{<eot>}$ token so we get back individual documents. At this point, these documents are already tokenized, so applying FIM at the token level is straightforward.

To transform data in the character space for context-level FIM, the tokenized documents have to be decoded back into strings before FIM augmentation. Depending on the vocabulary, some care has to be given to ensure decoding does not introduce any spurious characters into training. For example, utf-8 characters are encoded as multiple tokens with a BPE vocabulary; they can result in fragments from chunking and fail to decode. To prevent unforeseen errors midway through training, we encourage checking for these fragments at the beginning or end of a context and removing them.

After the transformed documents are encoded and joined back, the resulting context can be longer or shorter than the original, unaugmented context for context- and character-level FIM. For this reason, we recommend to trim or pad the transformed context to the model context length.

\section{Details of SPM encoding}\label{appendix:spm_details}
As mentioned in Section \ref{sec:fim_training}, in  SPM we use the ordering $[\text{suffix}, \text{prefix}, \text{middle}]$. In this section, we briefly discuss the choices regarding the sentinel tokens in SPM mode. A natural choice of encoding for SPM data would be to use
\[ \textsc{<suf>}\circ \text{Enc}(\text{suffix}) \circ \textsc{<pre>}  \circ \text{Enc}(\text{prefix})\circ \textsc{<mid>} \circ \text{Enc}(\text{middle}) \circ \textsc{<eot>}. \tag{SPM variant 1}
\] 
However, the encoding of SPM we use in this work is
\[
 \textsc{<pre>}  \circ \textsc{<suf>}\circ \text{Enc}(\text{suffix}) \circ \textsc{<mid>}\circ \text{Enc}(\text{prefix})\circ \text{Enc}(\text{middle}) \circ \textsc{<eot>}. \tag{SPM variant 2}
\]

The reason that we do not use the former is that it creates a separation between PSM and SPM, which may result to less transfer between SPM and PSM. To understand, note that with the second variant SPM data occurs naturally as part of PSM training since when we split a document uniformly at random, sometimes the chosen prefix will be empty. This is the reason pure PSM runs achieve strong performance when evaluated in SPM mode as in Table \ref{tab:pretraining:psm-spm}. 

Despite this, we note that the first SPM variant has its own advantages. In particular, it can be stronger in handling of subtokens at the end of prefix. Hence, the choice of which variant of SPM to use may depend on application in mind. As such, especially when training in pure SPM mode, it could be preferable to use the former simpler form. However, in this work, due to our emphasis on joint training of PSM and SPM and to maximize transfer between them, we opt for the second variant.
\section{Random span infilling benchmark}
\label{appendix:random-span-infilling}

\cite{incoder} introduced the single-line and multi-line infilling benchmarks based on HumanEval which prove valuable for measuring FIM performance. One limitation of these benchmarks is that the middle section is selected based on lines and does not capture more general use cases in the wild. We created a third infilling benchmark by choosing the middle span from two random  positions in the canonical solution. In this section, we show some examples of these tasks so the reader can get a feel for the new benchmark. The goal is to predict the highlighted span.

\vspace{-6pt}
\begin{center}
\begin{minipage}{5in}
\begin{lstlisting}[style=python]
from typing import List

def has_close_elements(numbers: List[float], threshold: float) -> bool:
    """ Check if in given list of numbers, are any two numbers closer to each other than
    given threshold.
    """
    for idx, elem in enumerate(numbers):
        for idx2, elem2 in enumerate(numbers):
            if idx != idx2:
              `  distance = abs(elem` - elem2)
                if distance < threshold:
                    return True

    return False
\end{lstlisting}
\end{minipage}
\end{center}

Here, for the model to pass, it needs to know that 1) the variable \texttt{distance} is not defined, 2) the prefix ends in a subtoken and not handling this will result in an indentation error, and 3) the completion has to stop in-line when the difference is calculated.

\vspace{-6pt}
\begin{center}
\begin{minipage}{5in}
\begin{lstlisting}[style=python]
def rounded_avg(n, m):
    """You are given two positive integers n and m, and your task is to compute the
    average of the integers from n through m (including n and m). 
    Round the answer to the nearest integer and convert that to binary.
    If n is greater than m, return -1.
    Example:
    rounded_avg(1, 5) => "0b11"
    rounded_avg(7, 5) => -1
    rounded_avg(10, 20) => "0b1111"
    rounded_avg(20, 33) => "0b11010"
    """
    if m < n:
        return -1
    summation = 0
    for i in range(n, m`+1):`
`        summation += i`
`    retu`rn bin(round(summation/(m - n + 1)))
\end{lstlisting}
\end{minipage}
\end{center}

This is a slightly more difficult example where the missing section spans over multiple lines and ends in a subtoken which would break all previous works that use BPE encoding and token-based FIM. Use cases like this can happen in coding assistants when the user does not like the current implementation and quickly deletes an approximate span they want replaced by a code model.

Because we create random span infilling tasks uniformly at random, this naturally captures problems of varying difficulties and corner cases that could happen in practice. We picked 10 random tasks per problem in HumanEval because 1640 tasks yielded a good balance between reducing evaluation variance and sampling time.

\begin{figure}[ht!]
\centering
    \begin{subfigure}[b]{\textwidth}
        \centering
        \includegraphics[width=\textwidth]{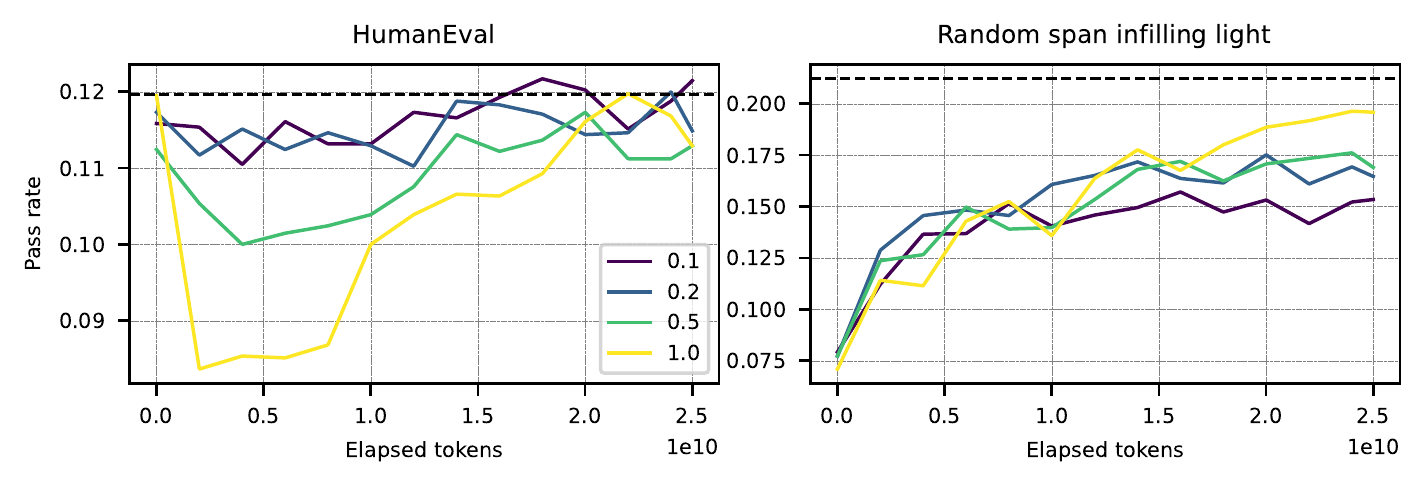}
        \caption{25B tokens of FIM finetuning.}
    \end{subfigure}
    \begin{subfigure}[b]{\textwidth}
        \centering
        \includegraphics[width=\textwidth]{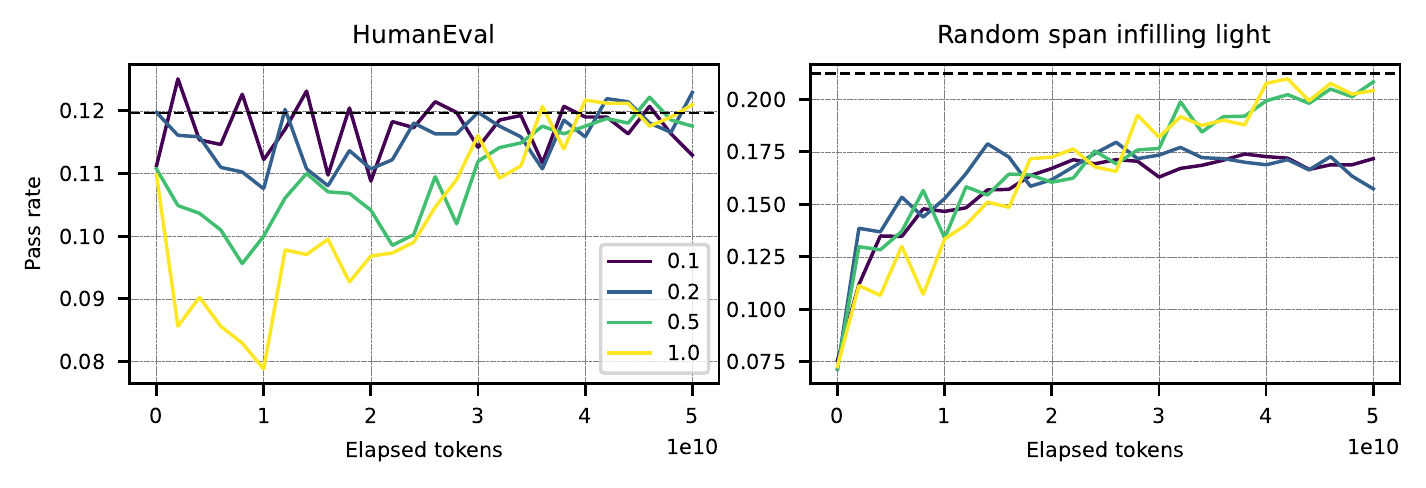}
        \caption{50B tokens of FIM finetuning.}
    \end{subfigure}
\caption{The dynamics of HumanEval and random span infilling light during finetuning. The legend corresponds to 
the fraction of finetuning
learning rate relative to the pretraining learning rate.  The results here are with a FIM rate of 0.9 and we omit similar dynamics plots with a FIM rate of 0.5 for brevity. } \label{fig:finetuning_dynamic}
\end{figure}

\section{Dynamics and learning curves of finetuning}\label{appendix:ft_dynamic}
To further build intuition about the results in Section \ref{sec:finetuning}, it is instructive to look at the dynamics of our infilling evaluation benchmarks during the finetuning. This is presented in Figure \ref{fig:finetuning_dynamic}. We observe that the ordinary HumanEval degrades significantly at the beginning of finetuning, especially  when using higher learning rates,  but it catches up to similar levels as pretraining by the end of the training. On the other hand, performance in random span infilling light starts out as zero as expected and slowly rises during finetuning.

\section{Top models comparison}\label{appendix:top_model}
In this section, we compare the performance of the current best infilling models on single-line, multi-line and random span infilling benchmarks. The results are reported in Table \ref{tab:top-models}.  We note that the numbers from InCoder in this table are self-reported numbers from the paper and was not independently evaluated in our framework. It is possible that minor differences in the implementation between our evaluation frameworks may result in slight discrepancies.

\begin{table}[ht!]
\centering

\begin{tabular}{cccc}
\hline
Model & Single-line infilling & Multi-line infilling & Random span infilling\\
\hline
\textsc{FIM50} & 0.730 & 0.406 & 0.521\\
\textsc{FIM90} &  0.751 & 0.441 &   0.551 \\
\textsc{InCoder} & 0.690 & 0.386 & N/A \\ \hline
\textsc{code-davinci-002} & 0.916 & 0.699 & 0.742  \\
\hline
\end{tabular}
\caption{Comparison of our 6.9B parameter (6.5B non-embedding parameters) FIM model  trained with a FIM rate of 50\% and 90\% for 100B tokens with the InCoder model of similar size (6.7B) and code-davinci-002, on the three main infilling benchmarks. All the FIM results are obtained in the SPM mode. We evaluated our models and code-davinci-002 using 100 samples our models and per task with a sampling temperature of 0.2.}
\label{tab:top-models}
\end{table}

\section{Qualitative evaluation}
\label{appendix:qual_eval}

Previously, we measured the pass rates on coding benchmarks to assess the infilling capability. In this section, we qualitatively evaluate samples to understand the strengths and areas of improvement for FIM. Overall, we find that infilling works better on the code domain than language. However, as previously motivated, infilling is generally a more difficult task than just extending a prefix. We exemplify these challenges and show possible mitigations.

\subsection{Successful infilling examples}
\label{sec:qual:good}

FIM enables a model to process information from both before and after the point of generation. This unlocks new capabilities that previously required specialized models finetuned on specific tasks. For example, unlike Codex \citep{codex} that trained a separate docstring model, we now have a single model that can infer the import modules, function names, arguments, docstrings, definitions, and many more. We show one such example below that is impossible to complete unless the model can read the entire source code. This example is also interesting in that the prefix ``\texttt{from sym}'' and the suffix both contain subtokens, which are known to cause traditional language models trained without techniques like stochastic BPE \citep{bpe-dropout} to fail.

\vspace{-6pt}
\begin{center}
\begin{minipage}{5in}
\begin{lstlisting}[style=python]
from sym`py import isprime`

`def largest_prime_factor(n):`
`    """`
`    Return the largest prime factor of n.`
`    "`""
    ans = 1
    for num in range(2, n + 1):
        if n % num == 0 and isprime(num):
            ans = num
    return ans
\end{lstlisting}
\end{minipage}
\end{center}

The benefits are not limited to coding. The model can adapt to the existing writing style and complete the passage in a natural way that takes the ending into consideration.

\vspace{-6pt}
\begin{center}
\begin{minipage}{5in}
\begin{lstlisting}[style=text]
Dolphins are` very intelligent animals. They are mammals and breathe air. They live in the sea and are related to whales and porpoises. Dolphins are very playful` animals.
\end{lstlisting}
\end{minipage}
\end{center}

\vspace{-18pt}
\begin{center}
\begin{minipage}{5in}
\begin{lstlisting}[style=text]
The commercial diver finally `thought he'd snagged a big catch when he saw something white. But then he quickly realized it wasn't a fish -- he` was wrangling an alligator.
\end{lstlisting}
\end{minipage}
\end{center}

\vspace{-18pt}
\begin{center}
\begin{minipage}{5in}
\begin{lstlisting}[style=text]
`Wikipedia is a free, web-based, collaborative, multilingual encyclopedia.` It is overseen by the nonprofit Wikimedia Foundation. Wikipedia uses a collaborative software known as wiki that facilitates the creation and development of articles.
\end{lstlisting}
\end{minipage}
\end{center}
\vspace{-3pt}

\subsection{Limitations}
\label{sec:qual:bad}

\textbf{Difficult prompts}. Unlike completing text from the end, infilling needs to infer the missing span that connects the prefix to the suffix. When the suffix is completely unrelated, the model can generate very long middle sections. We consider this behavior as the model's attempt at coming up with a plausible trajectory that joins the ending. Because the context size is limited, the model usually fails to join. However, given that even people have trouble infilling some of these prompts in a short passage, this failure demonstrates how challenging of a task FIM can be.

Below, we show one such difficult prompt where the model typically fails to connect entirely or join in a seamless way. Even when the model writes a seemingly plausible middle section, the quality can often vary.

\vspace{-6pt}
\begin{center}
\begin{minipage}{5in}
\begin{lstlisting}[style=text]
The dentist looked me in the eyes` and said, "I'm going to have to take all of your teeth out." I was stunned. I said, "All my teeth? Isn't there something else we could do?" He said, "No, I'm afraid not."`

`No one can predict the future.`

`The Ottomans were defeated in World War I` and the French were defeated at Waterloo.
\end{lstlisting}
\end{minipage}
\end{center}
\vspace{-3pt}


\textbf{Deciding when to stop}. The model is trained to predict the \textsc{<eot>} token when it thinks it has joined the suffix. Even when the prompts are seemingly straightforward, deciding when to end can still be a challenge in practice. Because there are many equally valid completions with varying lengths, the probability of outputting the \textsc{<eot>} is discounted by other longer candidates and is often smaller than expected. This is further exacerbated by the fact that the terminal symbol can simply be missed due to sampling. This results in a behavior where the model does not seem to end in a timely manner and generates a valid, but spurious content in the middle. In the process, the model can choose to write its own ending to the prefix, effectively ignoring the given suffix.

\vspace{-6pt}
\begin{center}
\begin{minipage}{5in}
\begin{lstlisting}[style=text]
Dogs are friendly animals.
Koalas are` pleasant animals.`
`Monkeys are playful animals.`
`Whales are enormous animals.`
`Owls are wise animals.`
`Penguins are graceful animals.`
`Crocodiles are ferocious` animals.
\end{lstlisting}
\end{minipage}
\end{center}
\vspace{-3pt}

While the general problem of not knowing when to stop applies to normal left-to-right completion as well, this has not been as big a problem as infilling because there is no constraint to join the suffix.

\textbf{Repetition}. When the model fails to generate an \textsc{<eot>} and copies the suffix, the model's ability to match patterns leads it to lock on and repeat the prompt indefinitely. Surprisingly, even large models are susceptible to this mode of failure. The example below ends with ``\textsf{\small the the heart,}'' because the model has failed to generate the terminal symbol and is still in the middle of filling in the missing span which unfortunately will not stop.

\vspace{-6pt}
\begin{center}
\begin{minipage}{5in}
\begin{lstlisting}[style=text]
The way is not in `the sky. The way is in the heart.`
`The way is not in the sky. The way is in the heart.`
`The way is not in the sky. The way is in the heart.`
`The way is not in the sky. The way is in the `the heart.
\end{lstlisting}
\end{minipage}
\end{center}
\vspace{-3pt}

\subsection{Mitigations}
\label{sec:qual:mitigations}

Like GPT-3 \citep{gpt3} where the performance depends on the quality of prompts, some of the failures in the earlier sections can be alleviated with prompt engineering. Namely, providing hints to constrain the output can dramatically improve the model's ability to generate the \textsc{<eot>} token and connect to the suffix within a reasonable token budget as the model has a more concrete understanding of how long the middle section should be.

One such idea is to provide examples both in the beginning and the end with numbered items. This makes the model internally keep track of the position, pay attention to the desired prefix and suffix, and generally abstain from generating spurious content as shown below. Providing leading examples alone without any explicit cues can often worsen the problem because it does not resolve the ambiguity in whether the model should join to the beginning of the suffix or consider it as part of a new example.

\vspace{-6pt}
\begin{center}
\begin{minipage}{5in}
\begin{lstlisting}[style=text]
1. Dogs are friendly animals.
2. Koalas are` sleepy` animals.
3. Lions are regal animals.
\end{lstlisting}
\end{minipage}
\end{center}

\vspace{-18pt}
\begin{center}
\begin{minipage}{5in}
\begin{lstlisting}[style=text]
Section 1:
1. The way is not in `the sky. The way is in `the heart.
2. Peace comes from within. Do not seek it without.
Section 2:
\end{lstlisting}
\end{minipage}
\end{center}
\vspace{-6pt}

It is important to note that the numbered few-shot prompting helps considerably but does not completely fix the problem, as the model can still accidentally start a new list of items.

In general, as the model can simply miss sampling the \textsc{<eot>} token, we recommend generating multiple samples and preferring samples that end with \textsc{<eot>}, as this increases the chance of choosing a sample that actually joins the ending. When multiple samples end in \textsc{<eot>}, they can be reranked by the likelihood or other heuristics of interest. We call this EOT-aware best-of-n sampling.

\end{document}

%% file: highlight.tex
\usepackage{listings}
\usepackage{xcolor}
\usepackage{atbegshi}
\usepackage{ifthen}

\usepackage{tikz}
\usetikzlibrary{calc}
\usetikzlibrary{fit}

\makeatletter

\newcommand\if@empty[1]{%
    \if\relax\detokenize{#1}\relax
        \expandafter\@firstoftwo
    \else
        \expandafter\@secondoftwo
    \fi
}

\newcommand\ifthen[2]{\ifthenelse{#1}{#2}{}}
\newcommand\ifelse[2]{\ifthenelse{#1}{}{#2}}

\newif\ifhl@active

\newtoks\hl@shipout

\newtoks\hl@parameters

\newcommand\hl[2][]{\begingroup\hlstyle@start{#1}#2\endgroup}

\newcommand\hlstyle[1][]{\hlstyle@start{#1}}
\newcommand\hlstyle@start[1]{%
    \global\hl@activetrue
    \aftergroup\hlstyle@end
    \write\@auxout{\noexpand\hl@setparameters{#1}}%
    \hl@mark{S}%
}
\newcommand\hlstyle@end{%
    \hl@mark{E}%
    \global\hl@activefalse
}

\newcommand\hl@setparameters[1]{%
    \global\hl@parameters={#1}%
}

\newcommand\hl@mark[1]{%
    \ensuremath{\vcenter{\hbox{\pdfsavepos}}}%
    \write\@auxout{\noexpand\hl@processmark{#1}{\the\pdflastxpos}{\the\pdflastypos}{\arabic{page}}}%
}

\AtBeginShipout{\AtBeginShipoutUpperLeft{%
    \pdfsavepos
    \write\@auxout{\noexpand\hl@processmark{Z}{\the\pdflastxpos}{\the\pdflastypos}{\arabic{page}}}%
    \hl@doshipout
}}

\newcommand\hl@processmark[4]{%
    \if#1S%
        \xdef\hl@firstx{#2}%
        \xdef\hl@firsty{#3}%
        \xdef\hl@firstp{#4}%
    \else\if#1M%
        \ifthenelse{\hl@firsty=#3}{%
            \xdef\hl@lastx{#2}%
            \xdef\hl@lasty{#3}%
        }{%
            \ifelse{\hl@firstx=\hl@lastx \and \hl@firsty=\hl@lasty}{%
                \expandafter\hl@draw\expandafter{\the\hl@parameters}%
            }%
            \xdef\hl@firstx{#2}%
            \xdef\hl@firsty{#3}%
            \xdef\hl@firstp{#4}%
        }%
    \else\if#1E%
        \ifthen{\hl@firsty=#3}{%
            \xdef\hl@lastx{#2}%
            \xdef\hl@lasty{#3}%
            \ifelse{\hl@firstx=#2}{%
                \expandafter\hl@draw\expandafter{\the\hl@parameters}%
            }%
        }%
    \else\if#1Z%
        \xdef\hl@zerox{#2}%
        \xdef\hl@zeroy{#3}%
    \fi\fi\fi\fi
}

\newcommand\hl@draw[1]{%
    \edef\@temp{{\noexpand\tikz [ overlay, shift = {(-\hl@zerox sp, -\hl@zeroy sp)} ]
        {\if@empty{#1}{\noexpand\hldefaultstyle}{\unexpanded{#1}}%
            {\hl@firstx sp}{\hl@firsty sp}{\hl@lastx sp}{\hl@lasty sp}}}%
        {\hl@firstp}%
    }%
    \expandafter\global\expandafter\addto@hook\expandafter\hl@shipout\expandafter{\@temp}%
}

\def\hl@doshipout{%
    \expandafter\hl@doshipout@\the\hl@shipout{}{}\@end
}
\def\hl@doshipout@#1#2#3\@end{%
    \if@empty{#2}{}{%
        \ifthen{\arabic{page}=#2}{#1}%
        \hl@doshipout@#3\@end
    }%
}

\let\orig@lst@discretionary=\lst@discretionary
\gdef\lst@discretionary{%
    \ifhl@active
        \hl@mark{M}%
    \fi
    \orig@lst@discretionary
    \ifhl@active
        \hl@mark{M}%
    \fi
}

\newcommand\hlboxstyle[5][red]{%
    \draw let \p1 = (#2, #3), \p2 = (#4, #5)
          in node [ draw = #1,
                    fill = #1!10!white,
                    minimum height = \baselineskip,
                    fit = (\p1)(\p2),
                    anchor = west,
                    align = none,  
                    outer sep = 0pt, inner sep = 0pt ]
             at (\p1) {};
}

\let\hldefaultstyle=\hlmarkerstyle

\makeatother

\newcommand\delimstyleA{\hlstyle[{\hlboxstyle[green]}]}

\newcommand\delimstyleBsm{\hlstyle[{\small\hlboxstyle[green!20, draw=green]}]}
\newcommand\delimstyleBft{\hlstyle[{\footnotesize\hlboxstyle[green!20, draw=green]}]}

\lstdefinestyle{python}
{
    columns=fullflexible,
    basicstyle={\ttfamily \small},
    moredelim=**[is][\delimstyleA]{@}{@},
    moredelim=**[is][\delimstyleBsm]{`}{`},
    keepspaces,
    breaklines,
    breakautoindent=false,
    breakindent=0pt,
}

\lstdefinestyle{text}
{
    columns=fullflexible,
    basicstyle={\sffamily \footnotesize},
    moredelim=**[is][\delimstyleA]{@}{@},
    moredelim=**[is][\delimstyleBft]{`}{`},
    keepspaces,
    breaklines,
    breakautoindent=false,
    breakindent=0pt,
}

%% file: doc-context-pass-rate-figure.tex
\begin{figure}[ht!]
\centering
\includegraphics[width=\textwidth]{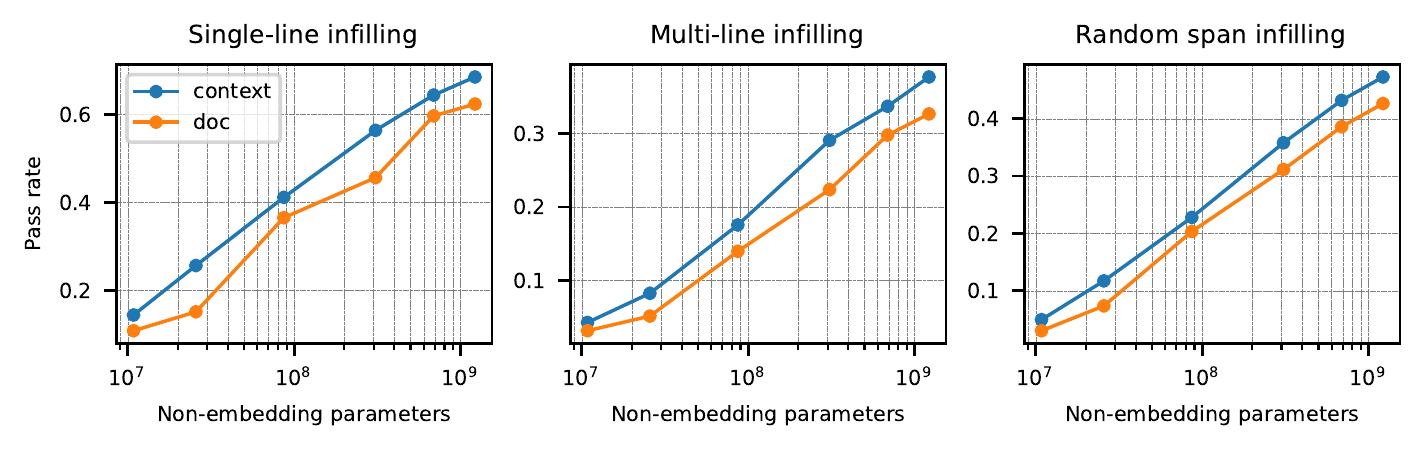}
\caption{
Applying FIM at the context level consistently outperforms document level FIM. All benchmarks are evaluated with temperature 0.2 and 200 samples/task.
}
\label{fig:pretraining:doc-context}
\end{figure}

%% file: doc-context-loss-figure.tex
\begin{figure}[ht!]
\centering
\includegraphics[width=\textwidth]{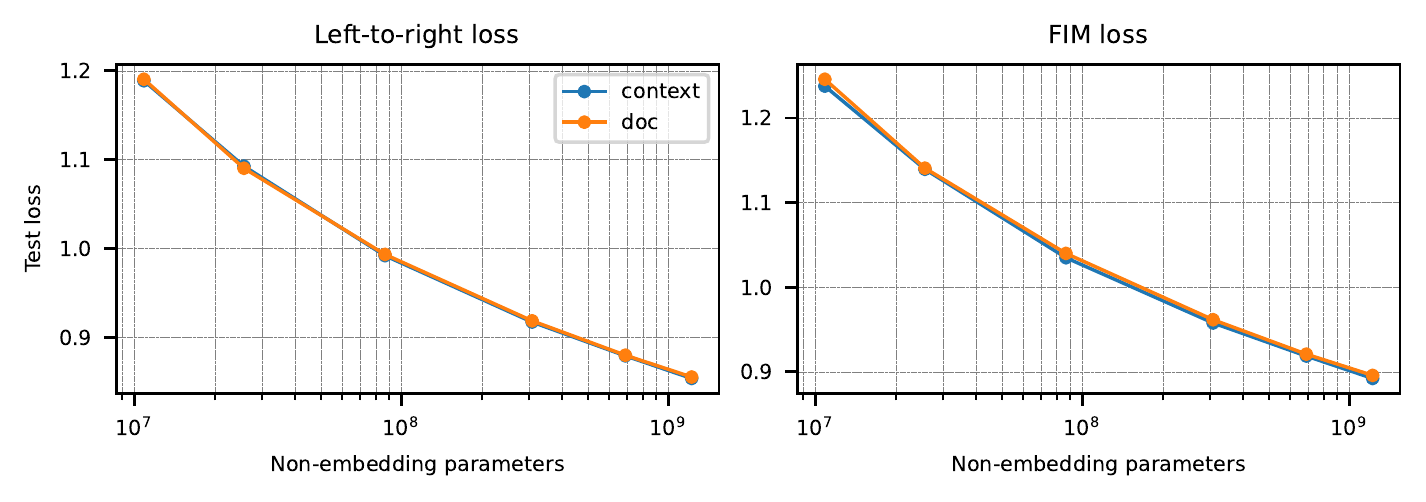}
\caption{
Comparison of losses with different FIM implementations. While document level FIM introduces partially broken data into training, it does not hurt the autoregressive loss (left). We also find that the reduction in FIM perplexity (right) is not commensurate to the gain in pass rate shown in Figure \ref{fig:pretraining:doc-context}.
}
\label{fig:pretraining:doc-context-loss}
\end{figure}